\RequirePackage{fix-cm}
\documentclass[twocolumn]{svjour3}          % twocolumn
\smartqed  % flush right qed marks, e.g. at end of proof

\usepackage{cite}
\usepackage{graphicx}
\usepackage{ragged2e}
\usepackage[tight,footnotesize]{subfigure}
\usepackage{rotating}
\usepackage{graphicx}
\usepackage{amsmath,amssymb} % define this before the line numbering.
\usepackage{array}
\usepackage{multirow}
\usepackage{colortbl}
\usepackage{amsfonts}
\usepackage{pifont}
\usepackage{xspace}
\usepackage{etoolbox}
\usepackage{overpic}
\usepackage{color}
\usepackage{microtype}

\usepackage{bm}
\usepackage{multirow}
\usepackage{color}
\usepackage{tabularx}
\usepackage{threeparttable}
\usepackage{bbding}
\usepackage{setspace}
\usepackage{booktabs}
\usepackage{xcolor}
\usepackage{colortbl}

\newcommand{\orcid}[1]{\href{https://orcid.org/#1}{\includegraphics[width=10pt]{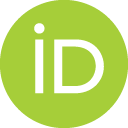}}}

\def\eg{\emph{e.g.}}
\def\ie{\emph{i.e.}}

\def\etal{{\em et al}}

\usepackage{hyperref}
\hypersetup{breaklinks=true,citecolor=blue, colorlinks}

\graphicspath{{./Imgs/}}
\DeclareGraphicsExtensions{.pdf,.jpg,.png}

\usepackage{silence}
\journalname{Research Article}

\begin{document}

\title{IBoxCLA: Towards Robust Box-supervised Segmentation of Polyp via Improved Box-dice and Contrastive Latent-anchors}

\titlerunning{IBoxCLA: Towards Robust Box-supervised Segmentation of Polyp}        % For running head

\author{Qiang Hu \orcid{0000-0001-8209-3754} \and 
  Ying Chen \orcid{0009-0007-1482-8246} \and 
  Hongkuan Shi \orcid{0000-0002-7655-5702} \and
  Qiang Li \orcid{0000-0002-9815-4432} \and \\
  Zhiwei Wang \orcid{0000-0002-1612-8573}
}

\authorrunning{Q. Hu \etal} % if too long for running head

\institute{
Qiang Hu, Ying Chen, Qiang Li, and Zhiwei Wang are with Wuhan National Laboratory for Optoelectronics, Huazhong University of Science and Technology, Wuhan, China.
(Email: {huqiang77, chenying2025, liqiang8, zwwang}@hust.edu.cn). \\
Hongkuan Shi is with Wuhan United Imaging Healthcare Surgical Technology Co.,Ltd, Wuhan, China. \\
Qiang Hu and Ying Chen are co-first authours.\\
Corresponding authors: Qiang Li and Zhiwei Wang.
% First Author and Second Author are with institution, (department), city, (state), country. 
% (Email: xxx@xxx.xx.xx, xxx@xxx.xx.xx). \\
% Third Author is with institution, (department), city, (state), country. 
% (Email: xxx@xxx.xx.xx). \\
% Corresponding author: XXXXXXX.
}

\date{Received: date / Accepted: date}
% The correct dates will be entered by the editor

\maketitle

% \begin{abstract}
% Please provide an abstract of 150 to 250 words. 
% The abstract should not contain any undefined abbreviations or 
% unspecified references.
% Please note: For some articles 
% (particularly, systematic reviews and original research articles), 
% 250 words may not be sufficient to provide all necessary information in the abstract. 
% Therefore, the abstract length can be increased from the 250-word limit 
% (to up to 450 words) if the topic dictates, 
% and to allow full compliance with the relevant reporting guidelines.

% % Please provide 4 to 6 keywords which can be used for indexing purposes.
% \keywords{First keyword \and Second keyword \and More}

% \end{abstract}
\begin{abstract}
Box-supervised polyp segmentation attracts increasing attention for its cost-effective potential. Existing solutions often rely on learning-free methods or pretrained models to laboriously generate pseudo masks, triggering Dice constraint subsequently. In this paper, we found that a model guided by the simplest box-filled masks can accurately predict polyp locations/sizes, but suffers from shape collapsing. In response, we propose two innovative learning fashions, Improved Box-dice (IBox) and Contrastive Latent-Anchors (CLA), and combine them to train a robust box-supervised model \emph{IBoxCLA}. The core idea behind IBoxCLA is to decouple the learning of location/size and shape, allowing for focused constraints on each of them. Specifically, IBox transforms the segmentation map into a proxy map using shape decoupling and confusion-region swapping sequentially. Within the proxy map, shapes are disentangled, while locations/sizes are encoded as box-like responses. By constraining the proxy map instead of the raw prediction, the box-filled mask can well supervise IBoxCLA without misleading its shape learning. Furthermore, CLA contributes to shape learning by generating two types of latent anchors, which are learned and updated using momentum and segmented polyps to steadily represent polyp and background features. The latent anchors facilitate IBoxCLA to capture discriminative features within and outside boxes in a contrastive manner, yielding clearer boundaries. We benchmark IBoxCLA on five public polyp datasets. The experimental results demonstrate the competitive performance of IBoxCLA compared to recent fully-supervised polyp segmentation methods, and its superiority over other box-supervised state-of-the-arts with a relative increase of overall mDice and mIoU by at least 6.5\% and 7.5\%, respectively. Codes will be released upon publication.

% Please provide 4 to 6 keywords which can be used for indexing purposes.

\end{abstract}
\section{Introduction}
\label{sec:introduction}
Colorectal cancer (CRC) is one of the most common cancers, seriously threatening human life and health~\cite{sung2021global}.
Polyps are early signs of CRC, and thus timely screening polyps by colonoscopy is of great importance for reducing the incidence and mortality~\cite{haggar2009colorectal}.
Since the doctors' skills and experience vary greatly, missing detection and misdiagnosis happen from time to time unfortunately.
Therefore, an automatic polyp segmentation method is of significant value to prompt doctors during colonoscopy and provide richer information for diagnosis.

During the past few decades, the fully-supervised methods of polyp segmentation \cite{fang2019selective,fan2020pranet,wei2021shallow,zhao2021automatic,dong2021polyp,zhang2022hsnet} have gained success but heavily rely on training samples with pixel-level annotations, which are prohibitively expensive. Moreover, boundaries between the polyps and normal tissues are often too indistinct to be delineated accurately even for experienced doctors. In comparison, it is more friendly and cheaper for doctors to only indicate polyps by drawing enclosed rectangles, \ie, boxes.
In view of this, a question may be asked: can we only build the polyps' box annotations, \eg, polyp detection datasets~\cite{ma2021ldpolypvideo,smedsrud2021kvasir,li2021colonoscopy}, to directly and robustly learn a polyp segmentor? Technically, the question could be answered using the concept of box-supervised segmentation.

\begin{figure}[t]
	\centering
	\includegraphics[width=\linewidth]{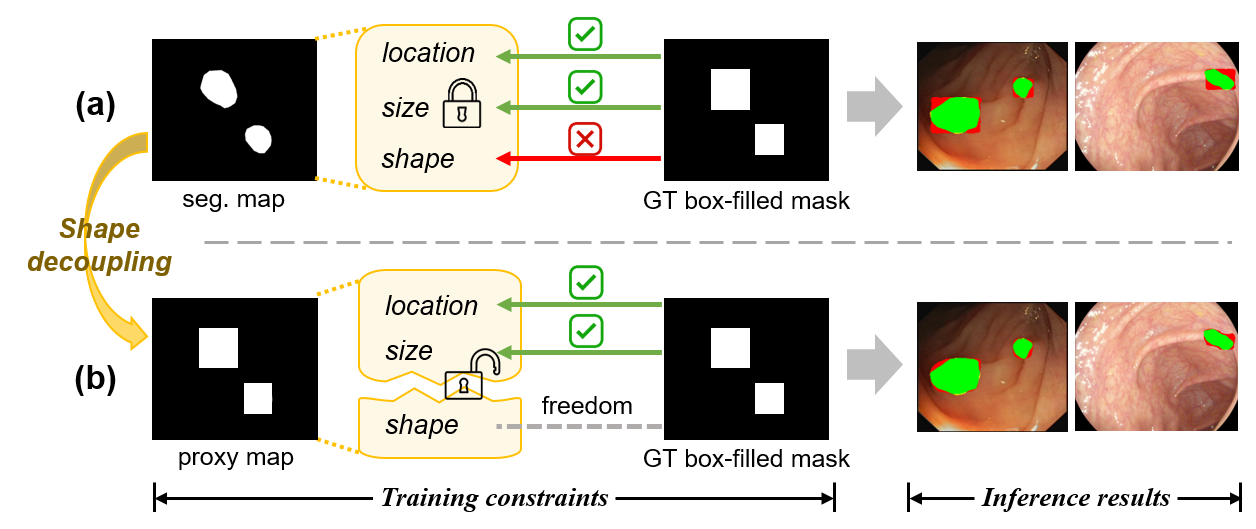}
	\caption{Comparison between (a) a simplistic method directly minimizing Dice loss between raw segmentation maps and box-filled masks and  (b) a prototype of our method, which applies Dice loss on the proxy maps of predictions, where the shapes are disentangled. Green: true positives; Red: false positives. As can be seen in the inference results, (a) the simplistic model is supervised by the incorrect shape information of the box-filled mask, and thus produces box-shaped over-segmentation results. In comparison, (b) the prototype of our method prevents itself from the misguiding supervision by using the shape-decoupled proxy maps, and thus can freely learn to segment boundaries.}
	\label{baseline}
\end{figure}

Box-supervised segmentation aims to leverage coarse box-level annotations to guide a model towards finer pixel-level prediction, a concept that has been extensively explored in nature scenes \cite{hsu2019weakly,tian2021boxinst,lan2021discobox,li2022box,cheng2023boxteacher}. One prevalent approach is to integrate box-supervised segmentation within the framework of multi-instance learning (MIL). In MIL constraint, each pixel is an instance, and positive bags are sampled from image regions intersecting with the ground-truth (GT) boxes, while negative bags are sampled from the non-overlapping regions. Bag-level predictions can be computed by max-pooling instance-level predictions within bags. 
Meanwhile, the prior of intensity inconsistency across boundaries is also coded as a complementary constraint like the color affinity loss in BoxInst~\cite{tian2021boxinst}.
% as a prir, but this is opposite to the characteristics of polyps in endoscopic images.
However, the polyps are often characterized by low color contrast, degrading the color affinity loss significantly. 
Besides, the bag sampling strategy varies in different MIL frameworks, often rendering the MIL-based methods complicated.

Another category of box-supervised segmentation methods offers a more intuitive alternative compared to the MIL-based approaches. These methods aim to convert box-supervised training into a standard fully-supervised learning paradigm by generating pseudo masks with the assistance of provided box annotations. Typically, they start by employing learning-free segmentation methods like GrabCut \cite{rother2004grabcut} and SLIC \cite{achanta2012slic}, or models pre-trained on other pixel-wise annotated images, to initialize pseudo masks based on the boxes, incorporating background information, color affinity, and tightness priors. The pseudo masks are then used to train a segmentation model with regular pixel-wise supervision like the Dice constraint.

Based on the above paradigm of mask generation, the idea of simply filling the boxes as pseudo masks triggers our curiosity. In this study, we train a segmentation model directly using the GT box-filled masks to supervise the predicted segmentation maps, as illustrated in Fig.~\ref{baseline}(a). Surprisingly, this simplistic method produces impressive results, accurately capturing polyp locations and sizes. The results inspire us to dig the potential of using the convenient and simple box-filled masks as supervisions, freeing us from building complex MIL frameworks or generating precise pseudo masks like the previous solutions. However, as can be seen in Fig.~\ref{baseline}(a), the simplistic method also suffers from over-segmentation, resulting in box-shaped predictions.
We argue that this phenomenon is caused by \emph{the entanglement of location/size and shape information in the raw segmented map}. Thus, directly optimizing the consistency between the segmented and box-filled masks can offer precise guidance to locations and sizes, but also introduce erroneous over-supervision on shapes.  Motivated by this, our aim in this paper is to eliminate this conflict by decoupling location/size and shape and ensure safe and robust supervision provided by GT box-filled masks, as illustrated in Fig.~\ref{baseline}(b).

To this end, we propose IBoxCLA, a box-supervised polyp segmentation method trained by two novel learning techniques: Improved Box-dice (IBox) and Contrastive Latent-Anchors (CLA). 
Specifically, IBox first constructs a proxy map for the segmentation map via shape decoupling and confusion-region swapping. During the decoupling, the segmentation map is squeezed into 1D orthogonal vectors, and then inversely projected into a 2D shape-decoupled proxy map. 
This decoupling process inevitably causes confusion regions between every two polyps. To address this, the proxy map is further rectified by identifying the potential confusion regions using the GT boxes, and swapping them with the corresponding raw predictions.
Then, IBox directly uses the GT box-filled masks as pixel-wise supervisions, but replaces the raw map with the proxy map in the Dice loss, thereby providing focused supervision for location/size learning while avoiding misleading shape constraints.

In addition to IBox, CLA plays a complementary yet crucial role in enhancing the shape boundaries by encouraging more discriminative feature learning. Inspired by the contrastive learning with momentum \cite{he2020momentum}, CLA creates an exponential moving average (EMA) replica of the original IBox-optimized model, serving as a teacher model. The prototypical feature anchors of polyp and background are computed by aggregating the teacher-extracted features.
CLA utilizes the two types of anchors to guide the original model in a contrastive manner. 
That is, outside the boxes, it forcefully pushes the features closer to the background anchor, facilitating the identification of intra-box background homogenous regions. 
Within the boxes, it pulls the highest
responding features on rows and columns further closer to the polyp anchor, effectively highlighting reliable polyp regions.
In summary, our contributions are listed as follows:
\begin{itemize}
 	\item[$\bullet$] We present IBoxCLA, a box-supervised polyp segmentation method that eliminates the need for complex MIL-derived frameworks and laborious pseudo mask generation. Instead, it offers an intuitive and robust approach towards precise pixel-wise predictions using simplistic box-filled masks as supervisions. 
	\item[$\bullet$] We propose two novel enabling learning techniques, namely IBox and CLA. IBox mitigates the negative impact of the incorrect shape information carried by the box-filled masks, enabling a focus on location and size learning. CLA enhances the feature contrast between polyps and backgrounds, facilitating more accurate predictions of shape boundaries.
	\item[$\bullet$] We benchmark our proposed IBoxCLA on five widely-used public datasets for polyp segmentation. The comprehensive experimental results demonstrate that IBoxCLA is comparable to or even better than the fully-supervised methods trained using the GT masks, and its superiority over the box-supervised state-of-the-arts with a relative increase of overall mDice and mIoU by at least 6.5\% and 7.5\%, respectively.
\end{itemize}
\section{Related Works}
\subsection{Fully-supervised Polyp Segmentation}
Most polyp segmentation methods \cite{fang2019selective,fan2020pranet,wei2021shallow,zhao2021automatic,dong2021polyp,zhang2022hsnet} rely on fully-supervised learning to train their model.
With the advent of U-Net~\cite{ronneberger2015u}, a plenty of U-Net variants \cite{zhou2018unet++,chen2021transunet,cao2022swin} has been proposed for general medical image segmentation.
Compared to them, the methods of polyp segmentation aim to develop task-specific designs by use of polyp characteristics.

For instance, PraNet~\cite{fan2020pranet} generated a global map as the initial guidance region and then used the reverse attention module to refine the boundary results.
For the large variance of size, MSNet~\cite{zhao2021automatic} designed a subtraction unit to produce distinguishable features between adjacent decoding layers, enhancing the robustness to the polyp size.
Polyp-PVT~\cite{dong2021polyp} utilized a transformer encoder, a cascaded fusion module, a camouflage identification module, and a similarity aggregation module, to effectively enhance the feature representations.
HSNet~\cite{zhang2022hsnet} designed a Transformer-CNN dual-branch structure, which combines long-term interactive features and neighboring details of polyps to improve polyp segmentation.

Despite their success, these methods require pixel-level annotations, which are difficult and time-consuming to obtain.
Moreover, the caused shortage of training samples may limit the performance in the clinical usage. 
Thus, it is of great significance in the clinical practice to explore cost-efficient approaches of polyp segmentation.

\subsection{MIL-based Box-supervised Segmentation}
To lower the annotation dependency, box-supervised segmentation methods have been extensively studied in both nature and medical domains.
Among them, the idea of multi-instance learning (MIL) is widely adopted in \cite{hsu2019weakly,tian2021boxinst,lan2021discobox,li2022box,wang2021bounding,cheng2023boxteacher}.
As the first endeavor, the work~\cite{hsu2019weakly} proposed to disassemble an image into bags which are essentially image rows and columns. The box annotation is then transformed into bag annotation by checking whether the bag and box intersect with each other, that is, the bag is positive if intersecting, and negative otherwise. The predictions in instance-level (pixel-level) are aggregated into bag-level using max-pooling. Therefore, an instance-level predictor can be trained eventually using a typical MIL paradigm.
Subsequently, different MIL-based variants~\cite{tian2021boxinst,wang2021bounding} were proposed and they differ mainly in terms of loss designs and bag sampling strategies.
For example, BoxInst~\cite{tian2021boxinst} proposed a projection loss to collectively constrain all row-bags or column-bags in 1D projected space by minimizing a 1D Dice loss.
The variant\cite{wang2021bounding} were built upon the prior work~\cite{hsu2019weakly}, and allows to sample bags from different angles instead of the original two fixed ones.

It is necessary to introduce auxiliary constraints since the MIL loss is often too weak.
To this end, BoxInst~\cite{tian2021boxinst} proposed a pairwise affinity loss, which encourages the adjacent pixels with high color similarity to be classified into the same category.
DiscoBox~\cite{lan2021discobox} enabled the model to learn the correspondence between intra-class objects across boxes, enhancing the semantic understanding of the model.
BoxTeacher~\cite{cheng2023boxteacher} created a teacher model to provide finer and confident guidance on the MIL-learned predictions.
Although the MIL concept successfully enables segmentation using only box annotations, it creates a mismatch between prediction and supervision, leading to over-complicated and non-intuitive bag-level constraints.

\subsection{Pseudo Mask for Box-supervised Segmentation}
Another group of box-supervised segmentation approaches aims to bridge the mismatch between pixel-level predictions and box-level annotations, and to directly use the segmentation constraints like Dice coefficient~\cite{milletari2016v} to train the model intuitively.
Some methods, for example BoxSup~\cite{dai2015boxsup}, Box2Seg~\cite{kulharia2020box2seg}, BoxShrink~\cite{groger2022boxshrink}, adopted traditional learning-free techniques, for example MCG~\cite{pont2016multiscale}, GrabCut~\cite{rother2004grabcut}, SLIC~\cite{achanta2012slic}, to produce initial pseudo masks, and introduced an iterative training algorithm and box annotations to refine the estimated masks.
To mitigate the error propagation in the iterative training, \cite{mahani2022bounding} used the entropy measure to obtain the uncertain region inside the GT box and reduced the supervised weights of the region.
There are also a few methods like \cite{chen2022weakly} and BoxPolyp~\cite{wei2022boxpolyp} generated pseudo masks by resorting to the segmentation models pre-trained using other pixel-level annotated images.

In spite of intuitiveness, both learning-free and pre-trained methods cause a non-trivial overhead for generating pseudo masks. 
In this work, we found that the simplistic pseudo masks generated by filling boxes can provide better guidance on the learning of polyp locations/sizes, but the shapes must be free to learn to avoid misguidance.
Moreover, some complementary constraints on shape learning are necessary, which is a consensus in the MIL-based methods. 

The most recent work, WeakPolyp~\cite{wei2023weakpolyp}, partially supported the above ideas.
They proposed a mask-to-box (M2B) operation to transform the segmented map into a box-shaped map, which likewise relaxes the shape misguidance.
However, our proposed IBoxCLA has two major improvements compared to WeakPolyp: 
i) M2B works for single-polyp images, while our IBox can handle more practical scenarios where multiple polyps often occur thanks to confusion-region swapping;
ii) WeakPolyp frees the shape learning but gives no other auxiliary constraints, while our CLA can further enhance the feature contrast between polyp and background regions, yielding more accurate boundary predictions.
\begin{figure*}[t]
	\centering
	\includegraphics[width=\linewidth]{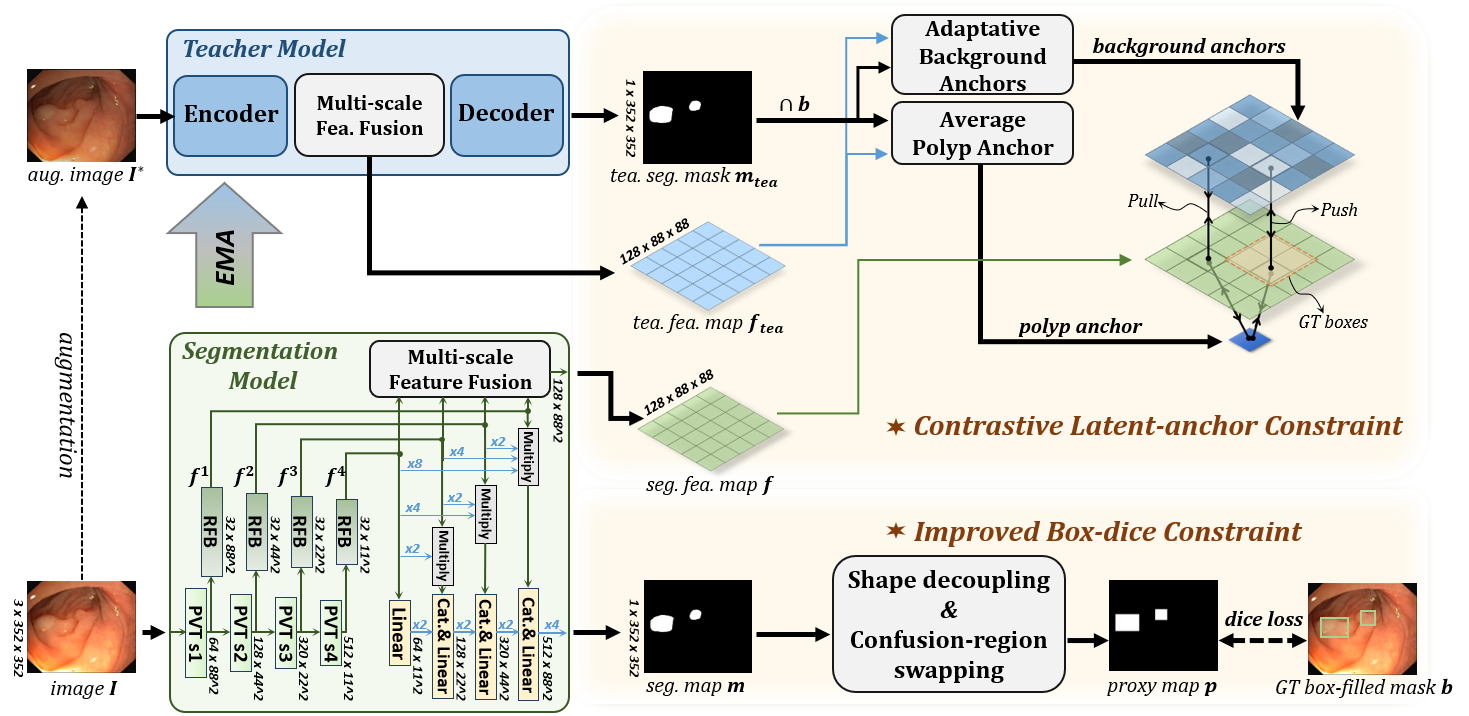}
	\caption{Illustration of the training phase of IBoxCLA. The segmentation model is for pixel-level predictions, and the teacher model is its EMA copy for providing latent anchors and masks. The segmentation model is constrained in three ways: (1) Improved Box-dice (IBox) constraint, (2) Contrastive Latent-Anchor (CLA) constraint, and (3) teacher-provided mask constraint (omitted in this figure).}
	\label{fig1}
\end{figure*}

\section{Methodology}
Fig.~\ref{fig1} illustrates the training phase of our proposed IBoxCLA, which consists of three major parts: (1) a segmentation model, (2) improved box-dice (IBox), and (3) contrastive latent-anchors (CLA).
In the following, we will detail each part and give implementation details.
\subsection{Segmentation Model for Pixel-level Prediction}

The green box in Fig.~\ref{fig1} gives details of our employed polyp segmentation model, which has three major parts, \ie, encoder, decoder and multi-scale feature fusion.
Given a 3-channel input image $\bm{I}$, we first resize it to a fixed size with $352 \times 352$ for the convenience of mini-batch gradient descent.

\subsubsection{Encoder}
The encoder contains a Pyramid Vision Transformer (PVT)~\cite{lee2022mpvit}, and four receptive field blocks (RFB)~\cite{wu2019cascaded}.
PVT utilizes four stages to down-scales the image size gradually, yielding four intermediate feature maps with the down-scaling factor of $4$, $8$, $16$, $32$, respectively, and the channel number of $64$, $128$, $320$, $512$, respectively.
In each stage, PVT first utilizes overlapping patch embedding to tokenize the output from the previous stage (or the raw image in the $1^{st}$ stage), and then introduces linear Spatial Reduction Attention (SRA) to efficiently process the vision-position embeddings, and lastly reshapes the embeddings into a feature map.

Four RFBs further reinvent the intermediate outputs of PVT into multi-scale feature maps, \ie, $\{\bm{f}^i \in \mathbb{R}^{32 \times {\frac{352}{2^{i+1}}} \times {\frac{352}{2^{i+1}}}}, i=1,2,3,4\}$. 
Each RFB has four branches with different respective fields, and each branch consists of a separable \cite{mamalet2012simplifying} and an atrous \cite{chen2017deeplab} convolution layer with the kernel/dilation size of $1$, $3$, $5$, $7$, respectively.
The outputs of the four branches are concatenated and linearly mapped to reduce the channel number to 32 by an additional $1\times1$ convolutional layer.

\subsubsection{Decoder}
%The decoder first transplants the information from the deeper layer(s) into each encoding feature map.
For $\bm{f}^i$, the decoder first up-scales the deeper features $\{\bm{f}^j, j=i,...,4\}$ into the same size with $\bm{f}^i$ using deconvolutional layers, and then multiplies them together as the updated $f^i$.
After that, the decoder progressively fuses the updated features using deconvolutional layers, concatenation operations, and linear mapping, sequentially, as indicated by the yellow boxes in Fig.~\ref{fig1}, yielding the final segmentation map, denoted as $\bm{m} \in \mathbb{R}^{1 \times 352 \times 352}$.

\subsubsection{Multi-scale Feature Fusion}
Multi-scale feature fusion constructs a feature space for the purpose of enhancing the feature contrast in the subsequent constraints.
Note that, the fusion process aims to derive useful gradients for the encoder, instead of providing features in the final segmentation.
Therefore, we adopt non-parametric operations for the multi-scale feature fusion.
Given the extracted $\{\bm{f}^i\}$, we simply up-scale them into the same size using bilinear interpolation, and concatenate them together to output a single feature map, denoted as $\bm{f} \in \mathbb{R}^{128 \times 88 \times 88}$.
% \subsection{Polyp-PVT}

\subsection{Improved Box-dice for Locations/Sizes Learning}
\label{method-Improved box-dice}
The GT polyp boxes are provided as coordinates indicating the top-left and bottom-right corners of each box.
We transform the GT coordinates into a binary GT mask $\bm{b} \in \mathbb{R}^{1 \times 352 \times 352}$ by assigning $1$ to the pixels within the boxes and $0$ outside them.

\begin{figure}[t]
	\centering
	\includegraphics[width=\linewidth]{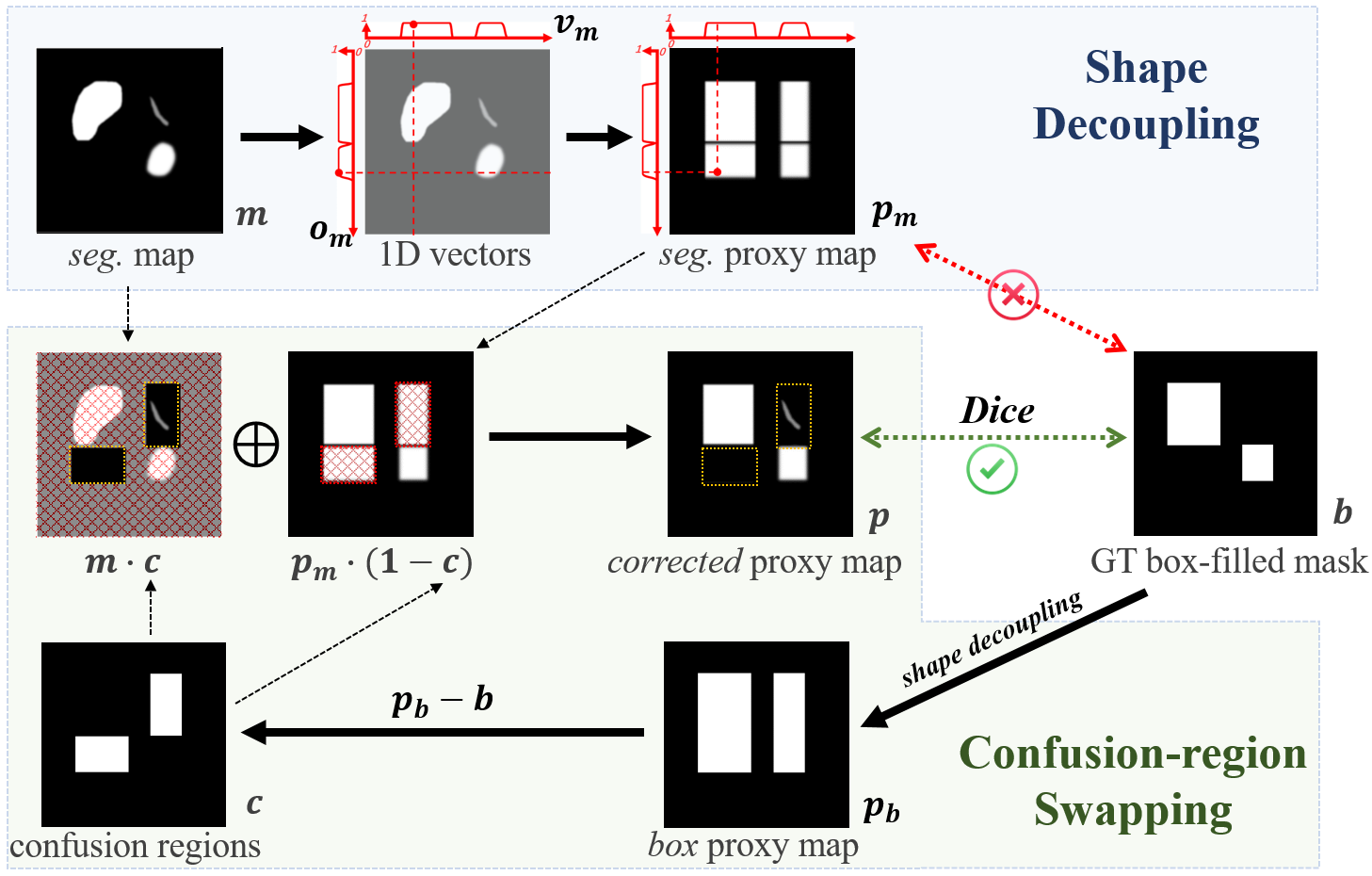}
	\caption{Details of IBox, consisting of shape decoupling and confusion-region swapping.}
	\label{fig2}
\end{figure}

Fig.~\ref{fig2} details our proposed IBox for constraining the above segmented map $\bm{m}$ using $\bm{b}$, and IBox has two major parts, \ie, shape decoupling and confusion-region swapping.

\subsubsection{Shape Decoupling}
As can be seen in Fig.~\ref{fig2}, the shape is inconsistent between $\bm{m}$ and $\bm{b}$, that is, $\bm{m}$ contains polyp boundaries, while $\bm{b}$ just implies the range of rows and columns where polyps occur in the image.
To eliminate the discrepancy, we first convert $\bm{m}$ into the 1D probability distributions of polyp along both row and column dimensions.

Specifically, as shown in Eq.~(\ref{eq1}), we apply 1D max-pooling on $\bm{m}$ horizontally and vertically, yielding two orthogonal vectors, \ie, $\bm{o_{m}}\in \mathbb{R}^{352 \times 1}$ and $\bm{v_{m}}\in \mathbb{R}^{1 \times 352}$, which align with the image width and height, respectively.
\begin{equation}
\bm{o_{m}}[i], \bm{v_{m}}[j] = \mathtt{max}(\bm{m}[i,:]), \mathtt{max}(\bm{m}[:,j]),
\label{eq1}
\end{equation}
where $\bm{m}[i,:]$ and $\bm{m}[:,j]$ are the $i^{th}$ row and $j^{th}$ column, respectively.
%where $\text{max}_{x}(\cdot)$ and $\text{max}_{y}(\cdot)$ indicate fetching the maximum values along the horizontal and vertical directions respectively.
Subsequently, the preliminary proxy map $\bm{p_{m}}$ is reconstructed to match the dimensions of $\bm{b}$ by inversely projecting $\bm{o_{m}}$ and $\bm{v_{m}}$ into a 2D space, which is formulated as $\bm{p_{m}} = \bm{o_{m}} \times \bm{v_{m}} \in \mathbb{R}^{1\times352 \times 352}$.
% follows: 
%\begin{equation} \small
%    p_{m}[i,j] = o_{m}[i] \times v_{m}[j], i=0,...,351, j=0,...,351.
%\end{equation}

If there is a single polyp, it is safe to directly calculate the dice coefficient between $\bm{p_{m}}$ and $\bm{b}$.
However, if there are multiply polyps, which is not rare, reconstruction errors occur during the inverse projection, invalidating the dice correspondingly, indicated by the red arrow in Fig.~\ref{fig2}.
This is because the 1D vectors not only remove the shape information, but also lose the separability, resulting in  false responses on the intersection of every two polyps.
We refer to these false regions in the proxy map as \emph{confusion regions}, and overcome this issue using confusion-region swapping.

\subsubsection{Confusion-region Swapping}
The confusion-regions are determined based on a simple fact that the GT box-filled mask should be identical before and after shape decoupling since it is originally shape-decoupled.
Therefore, as shown in Fig.~\ref{fig2}, we first apply the shape decoupling on $\bm{b}$, and then find the mismatched regions as the confusion-regions, denoted by $\bm{c}\in \mathbb{R}^{1\times352 \times 352}$.
That is, $\bm{c}=\bm{p_{b}}-\bm{b}$, where $\bm{p_{b}}=\bm{o_{b}}\times \bm{v_{b}}$ and $\bm{o_{b}}$ and $\bm{v_{b}}$ are two orthogonal vectors which are obtained by 1D max-pooling on $\bm{b}$.

False positives could occur in the regions indexed by non-zeros in $\bm{c}$.
To suppress them, we correct $\bm{p_{m}}$ into $\bm{p}$ by swapping values between $\bm{p_{m}}$ and $\bm{m}$ guided by $\bm{c}$.
We call this operation Confusion-region swapping, which is formulated as $\bm{p}= \bm{m}\cdot \bm{c} + \bm{p_{m}} \cdot (\bm{1}-\bm{c})$.
%n the confusion regionsthe  the  the GT box-filled Fig. (the orange regions in Fig. \ref{fig2}), denoted as $m_{con}$, and the formula is as follows:
%\begin{equation}
%    p= m\cdot c + p_{m} \cdot (1-c).
    % {\mathbbm{1}} \big( \sum_{i \neq j} max_{x} (Fill(B_i)) \cdot max_{y}(Fill(B_j)) \big)  \in \mathbb{R}^{H \times W},
%\end{equation}
%Then we swap the pixels between $m$ and $m_{dec}$ in confusion regions to get a proxy map $m_{pro}$, which is defined as follows:
% The optimized reconstructed mask  $m_{recon^*}$ is defined as follows: 
%\begin{equation} \small
%    m_{pro} =  m \circ m_{con} + m_{r} \circ (1-m_{con}).
%    \label{correct}
%\end{equation}
Finally, we compute and minimize the improved box-dice loss between $\bm{b}$ and $\bm{p}$, which is defined as follows: 
\begin{equation}
    \mathcal{L}_{ibox} = -\frac{2\times |\bm{b} \cap \bm{p}|}{|\bm{b}| + |\bm{p}|}.
    %= -\frac{2 \sum^{i=351}_{i=0}\sum^{j=351}_{j=0}(b[i,j] \cdot p[i,j])}{\sum^{i=351}_{i=0}\sum^{j=351}_{j=0}(b[i,j]+p[i,j])}.
    \label{IBox loss}
\end{equation}
%where $IBox(\cdot,\cdot)$ and $Dice(\cdot,\cdot)$ compute the Improved Box-dice loss and Sørensen-Dice coefficient between two 2-D maps, respectively.

\subsection{Contrastive Latent-anchors for Enhancing Shapes}
\label{method:CLA}
IBox frees the shape learning from the misguidance carried by $\bm{b}$, but is too weak to capture precise shape boundaries.
To address this, color affinity loss term is widely used in the MIL-based box-supervised methods~\cite{tian2021boxinst}, while it is unreliable in our case since polyps are flat and have low color contrast.
Inspired by the contrastive learning~\cite{he2020momentum}, we instead introduce CLA to directly enhance the discrimination between the learned features of polyp and background, and naturally expect to help produce clearer boundaries.

\subsubsection{Latent Anchor Generation}
The vanilla contrastive learning requires the paired inputs to constrain their feature distance, that is, minimizing the distance if the pair has the same class label or maximizing otherwise.
For $\bm{f}$ with the size of $88\times88$, there are approximately 30 million pairs of feature vectors, resulting in heavy computational cost.
To reduce the cost, we employ an anchor strategy derived from the famous triplet-loss~\cite{schroff2015facenet} as an equivalent to the above goal.

Instead of constructing massive pairs, we generate two types of anchors, one represents polyp class and the other represents background.
We thus pull the features within the polyp regions towards the polyp anchor, and push those from the non-polyp regions towards the background anchors.
Momentum~\cite{he2020momentum}, as one of the most effective techniques in contrastive learning, is also employed to help generate stable and robust latent anchors.
To this end, as shown in the top of Fig.~\ref{fig1}, we create a copy of the segmentation model named Teacher Model using exponential moving average (EMA), that is, $\bm{\theta_{tea}} = 0.99 \cdot \bm{\theta_{tea}} + 0.01 \cdot \bm{\theta_{seg}}$, where $\bm{\theta_{tea}}$ and $\bm{\theta_{seg}}$ are weights of the teacher and segmentation models, respectively.
The teacher model takes a perturbed copy of the image $\bm{I^{*}}$ as input, and produces a segmented map and a feature map, denoted as $\bm{m_{tea}} \in \mathbb{R}^{1\times 352 \times 352}$ and $\bm{f_{tea}} \in \mathbb{R}^{128\times 88 \times 88}$, respectively.

\begin{figure}[t]
	\centering
	\includegraphics[width=\linewidth]{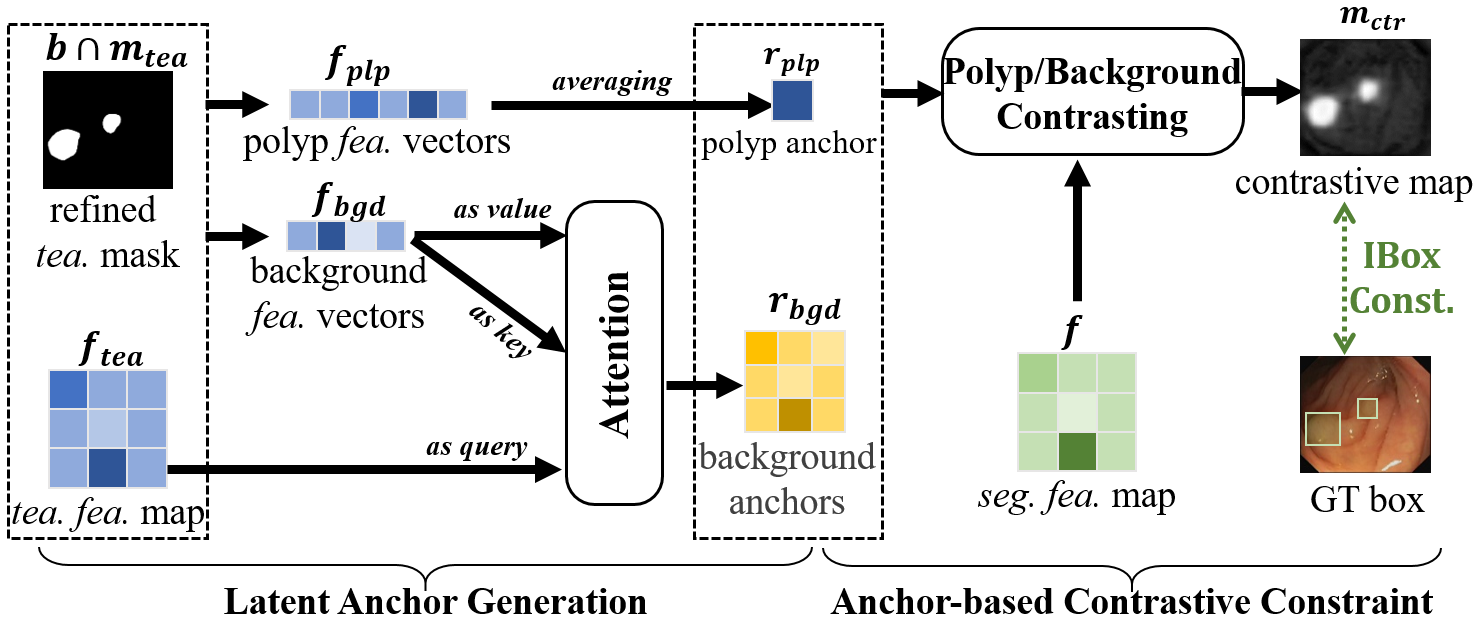}
	\caption{Details of CLA, consisting of latent anchor generation and anchor-based contrastive constraint.}
	\label{fig3}
\end{figure}
We notice that the latent anchors can be analogies with the prototypes in the few-shot learning task.
Therefore, we simply employ the advanced technique called Adaptive Background Prototype (ABP)~\cite{fan2022self} to generate our anchors. 
As shown in Fig.~\ref{fig3}, the polyp anchor $\bm{r_{plp}} \in \mathbb{R}^{128\times1 \times 1} $ is calculated by globally averaging all features from the polyp regions, while the background anchors $\bm{r_{bgd}} \in \mathbb{R}^{128\times88 \times 88} $ are generated for each position by dynamically aggregating similar background pixels using an attention mechanism.
The latent anchor generation is thus formulated as follows:
\begin{equation}
	\begin{split}
	\bm{r_{plp}} &= \frac{1}{|\bm{f_{plp}}|} \sum \bm{f_{plp}}, \\ 
	\bm{r_{bgd}} &= \bm{f_{bgd}} \times \mathtt{softmax}(\frac{\bm{f_{bgd}}^{T} \times \bm{f_{tea}} }{\sqrt{\bm{d_{k}}}}),
	\end{split}
\end{equation}
where $\bm{d_{k}}$ is a scale factor~\cite{vaswani2017attention}; $\bm{f_{plp}} \in \mathbb{R}^{128\times N} $ and $\bm{f_{bgd}}\in \mathbb{R}^{128\times M}$ are the feature vectors from $\bm{f_{tea}}$ indexed by the non-zero and zero regions in two versions of binarized $ (\bm{b} \cdot \bm{m_{tea}}) $ using high and low thresholds ($0.8$ and $0.5$, respectively) to discard vague regions on boundaries.

\subsubsection{Anchor-based Contrastive Constraint}
We first measure whether each feature in $\bm{f}$ is closer to the polyp or background anchors by calculating the cosine similarities and then using $\mathtt{softmax}$ to convert the similarities into probabilities, which is formulated as follows:
\begin{equation}
		\bm{m_{ctr}}[i,j] =  \frac{e^{\mathtt{cos}(\bm{f}[i,j],\bm{r_{plp}})}}{e^{\mathtt{cos}(\bm{f}[i,j],\bm{r_{plp}})}+e^{\mathtt{cos}(\bm{f}[i,j],\bm{r_{bgd}}[i,j])}}.
\end{equation}

We denote $\bm{m_{ctr}} \in \mathbb{R}^{1\times 88 \times 88} $ as contrastive map for polyp, where each entry represents the probability of being closer to polyp rather than background.
Our goal thus becomes suppressing $\bm{m_{ctr}}$ for true non-polyp pixels while highlighting it for true polyp regions.

However, the goal is ill-posed since only GT boxes are available.
To compromise, we only suppress the top false positives outside the GT boxes, and further enhance the top responses inside the GT boxes.
Fortunately, this can be easily accomplished using our proposed IBox constraint.
Therefore, we upscale $\bm{m_{ctr}}$ to the same size of $\bm{b}$, and the objective of CLA can be formulated as follows:
\begin{equation}
	\mathcal{L}_{cla} = -\frac{2 \times |\bm{b} \cap \bm{p_{ctr}}|}{|\bm{b}| + |\bm{p_{ctr}}|},
	%= -\frac{2 \sum^{i=351}_{i=0}\sum^{j=351}_{j=0}(b[i,j] \cdot p_{ctr}[i,j])}{\sum^{i=351}_{i=0}\sum^{j=351}_{j=0}(b[i,j]+p_{ctr}[i,j])},
\end{equation}
where $\bm{p_{ctr}}$ is the corrected proxy map of $\bm{m_{ctr}}$ obtained using shape decoupling and confusion-region swapping.

\subsection{Implementation Details}
\label{method-implement}
Besides $\mathcal{L}_{ibox}$ and $\mathcal{L}_{cla}$, we follow BoxTeacher~\cite{cheng2023boxteacher} and also utilize the teacher model to guide the pixel-level predictions, which is formulated as follows:
\begin{equation}
	\mathcal{L}_{px} = -\frac{2 \times |\bm{b} \cap \bm{m_{tea}} \cap \bm{m}|}{|\bm{b}\cap \bm{m_{tea}}| + |\bm{m}|}.
	\label{eq6}
\end{equation}

Therefore, the total loss can be calculated using Eq.~(\ref{eq7}).
\begin{equation}
	\mathcal{L}_{total} = \mathcal{L}_{ibox}+\mathcal{L}_{cla}+\mathcal{L}_{px}.
	\label{eq7}
\end{equation}

Note that, the gradients derived by minimizing $\mathcal{L}_{total}$ are utilized to update weights of the segmentation model only, while the teacher model is slowly and stably updated by aggregating historical weights of the segmentation model using EMA strategy.
Although both $\mathcal{L}_{cla}$ and $\mathcal{L}_{px}$ rely on the teacher model, they are not mutually exclusive.
$\mathcal{L}_{px}$ provides more explicit and efficient guidance, and $\mathcal{L}_{cla}$ constrains the feature learning which could be less vulnerable to the possible defective supervisions provided by the teacher model.

In implementations, we use PyTorch toolkit~\cite{paszke2019pytorch} to build our method, and a single NVIDIA RTX A6000 GPU to train it.
We adopt AdamW~\cite{loshchilov2017decoupled} as the optimizer, and set both the learning rate and the weight decay to $0.0001$.
We resize the input image into $352 \times352 $, and set the batch size and the number of epochs to $16$ and $100$, respectively.
For the teacher model, a random chromatic augmentation and a general multi-scale strategy widely employed in the previous works~\cite{fan2020pranet, wei2021shallow, zhao2021automatic, dong2021polyp,zhang2022hsnet} are utilized to generate the augmented input image.
\section{Experiments}
\subsection{Datasets and Evaluation Metrics}
\label{chapter-dataset}
\begin{table}[t]\small
	\centering
	\caption{Summary of polyp datasets used in this paper.}
    \scalebox{0.8}{
	\begin{threeparttable}
    \setlength{\tabcolsep}{3 pt}
	% \resizebox{\linewidth}{!}
		\begin{tabular}{lcccc}
			\toprule
			Datasets & Train Num. & Test Num. & Size & Anno. \\
			\midrule
			ClinicDB & \cellcolor{red!30}550 & \cellcolor{green!30}62 & $384\times288$ & Mask \\
			Kvasir-SEG & \cellcolor{red!30}900 & \cellcolor{green!30}100 & ($332$-$1920$)$\times$($487$-$1072$) & Mask\\
			ColonDB& - & \cellcolor{green!30}380 & $574\times500$ & Mask\\
			EndoScene & - & \cellcolor{green!30}60 & $574\times500$ & Mask \\
			ETIS& - & \cellcolor{green!30}196 &$1225\times966$ & Mask\\
			%\midrule
			LDPolypVideo & \cellcolor{yellow}33,884 & - & $560\times480$ & Box\\
			\bottomrule
		\end{tabular}
	\begin{tablenotes}
		\item[] The red, yellow and green marks indicate the original train-set, the extra train-set  and the test-set, respectively.
	\end{tablenotes}
\end{threeparttable}
}
\label{table0}
\end{table}

We conduct the experiments using six mostly-used public polyp datasets, \ie, ClinicDB \cite{bernal2015wm}, Kvasir-SEG \cite{jha2020kvasir}, ColonDB \cite{tajbakhsh2015automated}, EndoScene \cite{vazquez2017benchmark}, ETIS \cite{silva2014toward}, and LDPolypVideo~\cite{ma2021ldpolypvideo}, which are summarized in Table~\ref{table0}.
To be consistent with the previous works~\cite{fan2020pranet,wei2021shallow,zhao2021automatic,dong2021polyp,zhang2022hsnet}, a total of 1,450 samples from ClinicDB and Kvasir-SEG are viewd as an \emph{original} train-set.
LDPolypVideo is treated as an \emph{extra} train-set.
The rest 798 samples are used as test-set.
The annotated masks of the original train-set are converted into GT boxes by finding circumscribed boxes for connected components.

For the convenience of comparison, we employ Dice~\cite{milletari2016v} and IoU~\cite{yu2016unitbox} as the evaluation metrics, commonly utilized in the previous methods~\cite{fan2020pranet,wei2021shallow,zhao2021automatic,dong2021polyp,zhang2022hsnet,wei2023weakpolyp,wei2022boxpolyp}.
Specifically, for each dataset, we report mDice and mIoU, which are the average value at 256 different binarization thresholds from $0$ to $1$.

To verify the performance for location/size and shape separately, we split the segmentation task into detection and delineation sub-tasks in the evaluation.
Specifically, the segmentation results are converted into individual connected component regions.
For the detection sub-task, precision (P), recall (R) and F1-score (F1) are calculated by checking the IoU between the regions' enclosed boxes and the GT boxes (true positive if IoU $>0.5$).
For the delineation sub-task, Hausdoff distance (HD) between each true positive region and its corresponding mask is computed and averaged for all true positives.

\subsection{Comparison Methods}
We include fourteen state-of-the-art (SOTA) methods for comparison, including nine box-supervised methods~\cite{tian2021boxinst,lan2021discobox,li2022box,cheng2023boxteacher,groger2022boxshrink,mahani2022bounding,wang2021bounding,wei2023weakpolyp,wei2022boxpolyp}, and five fully-supervised~\cite{fan2020pranet,wei2021shallow,zhao2021automatic,dong2021polyp,zhang2022hsnet}.

For the box-supervised methods, BoxInst~\cite{tian2021boxinst}, DiscoBox~\cite{lan2021discobox}, BoxLevelSet~\cite{li2022box}, BoxTeacher~\cite{cheng2023boxteacher}, \cite{mahani2022bounding} and~\cite{wang2021bounding} were for natural images or other lesions.
Boxshrink~\cite{groger2022boxshrink} and WeakPolyp~\cite{wei2023weakpolyp} did not report the results on the popular datasets listed in Table~\ref{table0}.
Fortunately, these methods are all open source.
Thus, we use their released models with default parameters for comparison.
The box-supervised methods are all trained using the images from the original train-set and their converted boxes.
For the fully-supervised methods, \ie, PraNet \cite{fan2020pranet}, SANet \cite{wei2021shallow}, MSNet \cite{zhao2021automatic}, Polyp-PVT \cite{dong2021polyp} and HSNet \cite{zhang2022hsnet}, we copy the reported results from their papers for comparison.

In addition, we also extend several box-supervised methods into mix-supervised ones.
Inspired by BoxPolyp~\cite{wei2022boxpolyp}, the original train-set provides the regular mask supervisions and LDPolypVideo is added together to additionally provide box supervisions.
We treat these trained box-supervised methods as their mix-supervised counterparts, and it is more practically meaningful to compare them with the fully-supervised methods since the greatest advantage of box-supervised learning is \emph{the capability of increasing data amount with cheap annotation cost}.

It is worth noting that for every comparison in the following, we perform the paired-samples t-test~\cite{student1908probable} on the statement and report the $p$-values, which less than or equal to 0.05 were considered statistically significant.

\subsection{Comparison with Box-supervised SOTAs}
% \subsubsection{Training with only box annotations}
\label{com_box}
\begin{table*}[t]\small
	\centering
	\caption{Comparison segmentation results between IBoxCLA and other box-supervised methods.}
	%\resizebox{\linewidth}{44mm}{
		\scalebox{0.8}{
			\begin{threeparttable}
				\begin{tabular}{lccccccccccccc}
					\toprule
					\multirow{2}{*}{Methods}& \multirow{2}{*}{Supervision}& \multicolumn{2}{c}{ClinicDB}  & \multicolumn{2}{c}{Kvasir-SEG} & \multicolumn{2}{c}{ColonDB} & \multicolumn{2}{c}{EndoScene} & \multicolumn{2}{c}{ETIS}  & \multicolumn{2}{c}{Overall}\\
					& & mDice &mIoU & mDice &mIoU & mDice &mIoU & mDice &mIoU & mDice &mIoU & mDice &mIoU \\ 
					\midrule
					% \em{box-supervised methods:}& & & & & & & & & &\\
					Boxshrink~\cite{groger2022boxshrink} &Box &0.791 &0.668 &0.828 &0.728 &0.685 &0.559 &0.770 &0.641 &0.633 &0.504 &0.705 &0.581\\
					Mahani~\emph{et al.}~\cite{mahani2022bounding} &Box &0.812 &0.708 &0.811 &0.694 &0.716 &0.595 &0.827 &0.725 &0.659 &0.541 &0.730 &0.613\\
					Wang~\emph{et al.} \cite{wang2021bounding} &Box &0.840 &0.749 &0.863 &0.788 &0.718 &0.609 &0.832 &0.686 &0.668 &0.576 &0.742 &0.640\\
					BoxInst~\cite{tian2021boxinst} &Box &0.722 &0.632 &0.721 &0.636 &0.692 &0.599 &0.771 &0.684 &0.607 &0.525 &0.683 &0.594\\
					DiscoBox~\cite{lan2021discobox}  &Box &0.848 &0.773 &0.841 &0.759 &0.702 &0.608 &0.809 &0.726 &0.633 &0.547 &0.722 &0.634\\
					BoxLevelSet~\cite{li2022box} &Box &{0.888} &{0.823} &0.855 &0.785 &0.685 &0.596 &0.838 &0.764 &0.647 &0.564 &0.724 &0.642\\
					BoxTeacher~\cite{cheng2023boxteacher} &Box &0.779 &0.713 &0.854 &0.782 &0.730 &0.652 &0.846 &0.778 &0.582 &0.500 &0.722 &0.645\\
					WeakPolyp~\cite{wei2023weakpolyp} &Box &0.863 &0.794 &{0.878} &{0.815} &{0.766} &{0.676} &{0.866} &{0.790} &{0.678} &{0.604} &{0.774} &{0.694}\\ \rowcolor{gray!25}
					IBoxCLA (Ours) &Box &\textbf{0.916}$^{\dagger}$ &\textbf{0.858}$^{\dagger}$ &\textbf{0.907}$^{\dagger}$ &\textbf{0.851}$^{\dagger}$ &\textbf{0.802}$^{\dagger}$ &\textbf{0.716}$^{\dagger}$ &\textbf{0.882}$^{\dagger}$ &\textbf{0.813}$^{\dagger}$ &\textbf{0.779}$^{\dagger}$ &\textbf{0.696}$^{\dagger}$ &\textbf{0.824}$^{\dagger}$ &\textbf{0.746}$^{\dagger}$\\
					\bottomrule
				\end{tabular}
				\begin{tablenotes}
					\item[] $\dagger$ means the difference between IBoxCLA and the second-best WeakPolyp~\cite{wei2023weakpolyp} is statistically significant ($p<0.05$).
				\end{tablenotes}
			\end{threeparttable}
			\label{table1}
		}
	\end{table*}

\begin{figure*}[t]
	\centering
	\includegraphics[width=\linewidth]{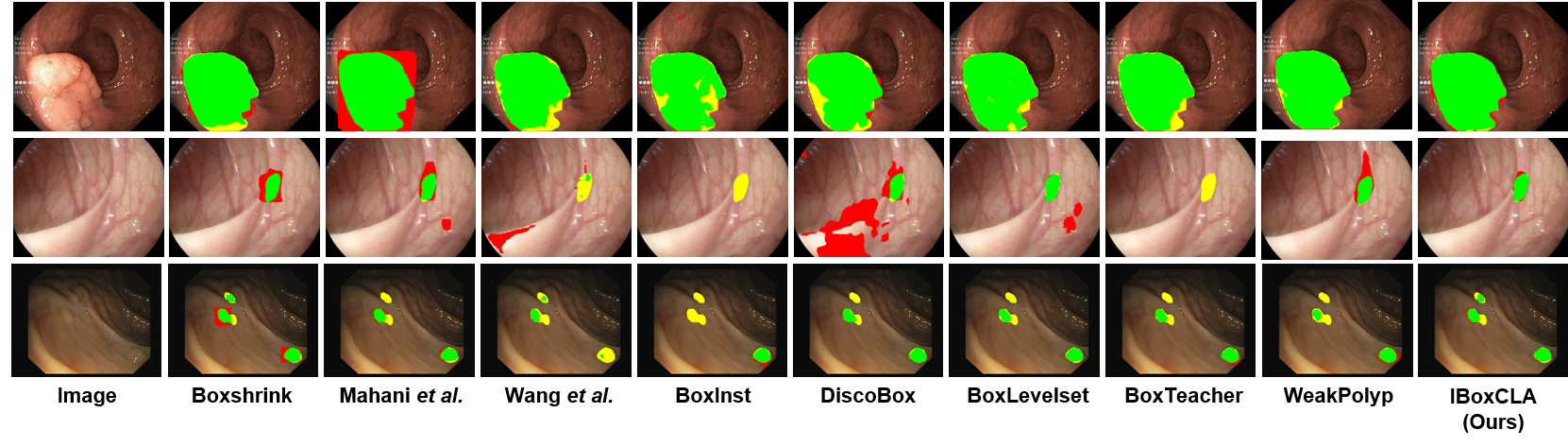}
	\caption{Three visualization results of different box-supervised methods. Green: true positives; Yellow: false negatives; Red: false positives.}
	\label{box-supervised methods visual}
\end{figure*}

Table~\ref{table1} presents the comparison results between IBoxCLA and the box-supervised SOTAs.
As can be seen, IBoxCLA outperforms other box-supervised methods by a significant margin both on samples from each dataset and on overall samples.
Moreover, IBoxCLA surpasses the second-best method, WeakPolyp, by $6.5\%$ and  $7.5\%$ in terms of mDice and mIoU on overall samples, respectively ($p<0.001$ for both metrics).
This is because that WeakPolyp ignores the confusion regions in shape decoupling and complementary constraints in shape learning, while our proposed IBox and CLA successfully address the two key issues, respectively.

Fig.~\ref{box-supervised methods visual} shows the segmentation results of three samples by different methods.
The methods relying on pre-generated pseudo masks are listed at the $2^{nd}$ to $3^{rd}$ columns.
They can roughly localize the polyps but fail to predict accurate boundaries.
The methods from the $4^{th}$ to $8^{th}$ columns are all MIL-based approaches and mostly resort to low-level intensity relationships for capturing shapes.
As can be seen, such relationships are too weak for flat and subtle polyps, resulting in non-trivial missing and false positive predictions.
Similar to ours, WeakPolyp also uses the box-filled masks to directly train a segmentation model, but it ignores the misguidance caused by confusion regions and the importance of auxiliary shape constraints.
In comparison, IBoxCLA recalls all polyps with clearer boundaries.

\subsection{Comparison with SOTAs in Practical Settings}
In clinical practice, the user has to annotate images at pixel-level to train a segmentor using fully-supervised learning.
One of our motivations is to let the model enjoy more data with a little cost of easy and quick box-level annotations.
Thus, we make a comparison between the fully-supervised and mix-supervised methods.
Specifically, WeakPolyp~\cite{wei2023weakpolyp} (the runner-up in Table~\ref{table1}), is extended to a mix-supervised version and included in the comparison.
BoxPolyp~\cite{wei2022boxpolyp}, as an original mix-supervised method, is also included.

For IBoxCLA, we train two variants in the fully-supervised and mix-supervised manners, respectively.
We disable the three losses in Eq.~(\ref{eq7}), and only use BCE loss plus Dice loss to train the pure segmentation model as our fully-supervised variant.
The quantitative comparison results are listed in Table~\ref{table2}.

\label{com_fully}
\begin{table*}[t]\small	
    \centering
    \caption{Comparison segmentation results between IBoxCLA and the fully-supervised and mix-supervised methods.}
    % \resizebox{\linewidth}{0.45\textwidth}{
    \scalebox{0.8}{
    	\begin{threeparttable}
    \begin{tabular}{lccccccccccccc}
    \toprule
    \multirow{2}{*}{Methods}& \multirow{2}{*}{Supervision}& \multicolumn{2}{c}{ClinicDB}  & \multicolumn{2}{c}{Kvasir-SEG} & \multicolumn{2}{c}{ColonDB} & \multicolumn{2}{c}{EndoScene} & \multicolumn{2}{c}{ETIS} & \multicolumn{2}{c}{Overall}  \\
		& & mDice &mIoU & mDice &mIoU & mDice &mIoU & mDice &mIoU & mDice &mIoU & mDice &mIoU \\
        \midrule
		% \em{fully-supervised methods:}& & & & & & & & & &\\
        % \rowcolor{gray!30} polyp-mask sup. &0.937 &0.889 &0.917 &0.864 &0.808 &0.727 &0.900 &0.833 &0.787 &0.706\\
            PraNet~\cite{fan2020pranet} &Mask &0.899 &0.849 &0.898 &0.840 &0.712 &0.640 &0.871 &0.797 &0.628 &0.567 &0.741 &0.675\\
            SANet~\cite{wei2021shallow} &Mask &0.916 &0.889 &0.904 &0.847 &0.753 &0.670 &0.888 &0.815 &0.750 &0.654 &0.794 &0.716\\
            MSNet~\cite{zhao2021automatic} &Mask &0.921 &0.879 &0.907 &0.862 &0.755 &0.678 &0.869 &0.807 &0.719 &0.664 &0.796 &0.717\\
            Polyp-PVT~\cite{dong2021polyp} &Mask &{0.937} &{0.889} &0.917 &0.864 &0.808 &{0.727} &{0.900} &{0.833} &0.787 &0.706 &0.833 &{0.760}\\
            HSNet~\cite{zhang2022hsnet} &Mask &\textbf{0.948} &\textbf{0.905} &{0.926} &{0.877} &{0.810} &{0.735} &{0.903} &{0.833} &{0.839} &{0.734} &{0.849} &{0.773}\\ \rowcolor{gray!25}
            IBoxCLA(Ours) &Mask
            &0.933 &0.887 &{0.921} &{0.869} &{0.809} &0.725 &0.875 &{0.809} &{0.793} &{0.712} &{0.833} &0.758\\
            \midrule
            WeakPolyp~\cite{wei2023weakpolyp} &Mix &{0.925} &{0.876} &{0.923} &0.865 &0.815 &0.734 &0.900 &0.827 &0.822 &0.734 &0.845 &0.768\\
            BoxPolyp~\cite{wei2022boxpolyp} &Mix &0.918 &0.868 &0.918 &{0.868} &{0.819} &{0.739} &{0.906} &{0.840} &{0.842} &{0.755} &{0.851} &{0.776}\\ \rowcolor{gray!25}
            IBoxCLA(Ours) &Mix &{0.937}$^{\dagger}$ &{0.892}$^{\dagger}$ &\textbf{0.927}$^{\dagger}$ &\textbf{0.878}$^{\dagger}$ &\textbf{0.848}$^{\dagger}$ &\textbf{0.769}$^{\dagger}$ &\textbf{0.915}$^{\dagger}$ &\textbf{0.851}$^{\dagger}$ &\textbf{0.846}$^{\dagger}$ &\textbf{0.767}$^{\dagger}$ &\textbf{0.869}$^{\dagger}$ &\textbf{0.798}$^{\dagger}$\\
    \bottomrule
    \end{tabular}
\begin{tablenotes}
	\item[] `Mask' means using the mask annotations from the original train-set to train models; `Mix' means using both the masks from the original train-set and the boxes from the extra train-set for training. $\dagger$ means the difference between IBoxCLA (Mix) and the second-best BoxPolyp (Mix)~\cite{wei2022boxpolyp} is statistically significant ($p<0.05$).
\end{tablenotes}
\end{threeparttable}
    \label{table2}
    }
\end{table*}

\begin{figure*}[t]
	\centering
	\includegraphics[width=0.98\linewidth]{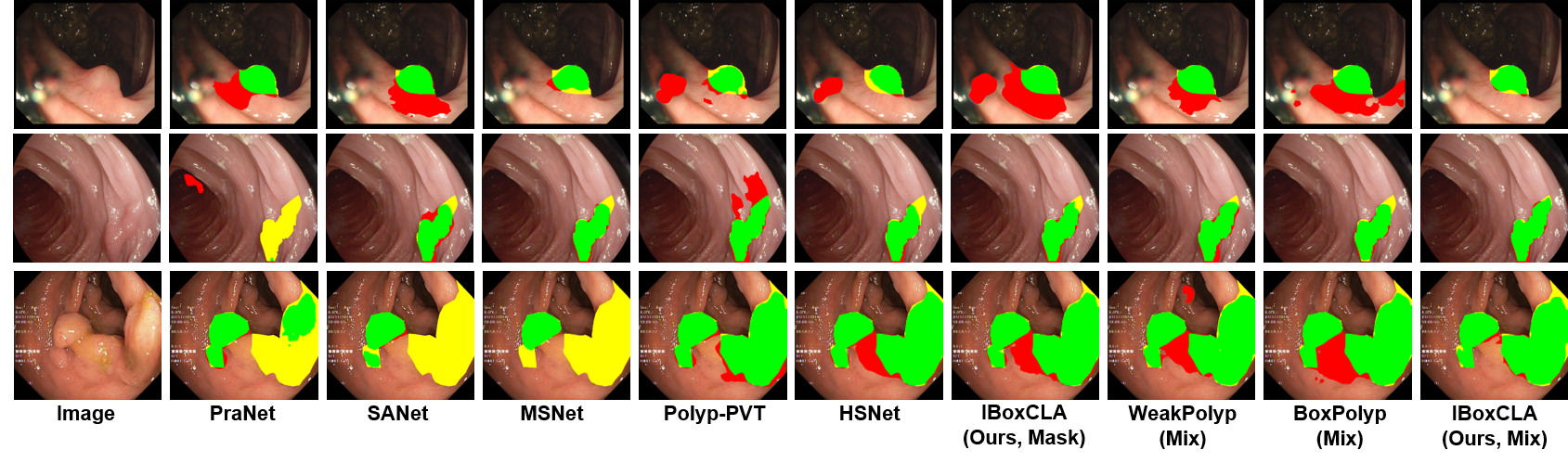}
	\caption{Three visualization results of different fully- and mix-supervised methods. Green: true positives; Yellow: false negatives; Red: false positives.}
	\label{fig4}
\end{figure*}

As can be seen, IBoxCLA (Mix) outperforms both the fully-supervised method, and the other mix-supervised methods.
This well verifies the practical value of our work and higher efficiency of absorbing box-guided knowledge from other datasets, \ie, LDPolypVideo in our case.
More in-depth analysis of learning efficiency is presented in Sec.~\ref{sec. learning eff}.

Moreover, all mix-supervised methods are close to or even surpass the fully-supervised methods.
For example, IBoxCLA (Mix) gains a significant improvement compared to IBoxCLA (Mask), and easily exceeds HSNet by an increase of mDice and mIoU on overall samples by $2.4\%$ and $3.2\%$, respectively ($p=0.002$, $p=0.006$).
This implies that using cheaper labels is a promising solution to further boost the performance  compared to developing more sophistic network architectures.

The fully-supervised methods all release their trained models, and thus we also make a qualitative comparison in Fig.~\ref{fig4}.
It can be clearly seen that most methods including IBoxCLA (Mask) are easily distracted by the polyp-surrounding tissues, resulting in severe over- and under-segmentation.
In comparison, IBoxCLA (Mix) predicts the best visual results with precise boundaries.

\subsection{Ablation Study}
\subsubsection{Effectiveness of the three loss terms}
\label{cha-ablation}
\begin{figure}[t]
	\centering
	\includegraphics[width=\linewidth]{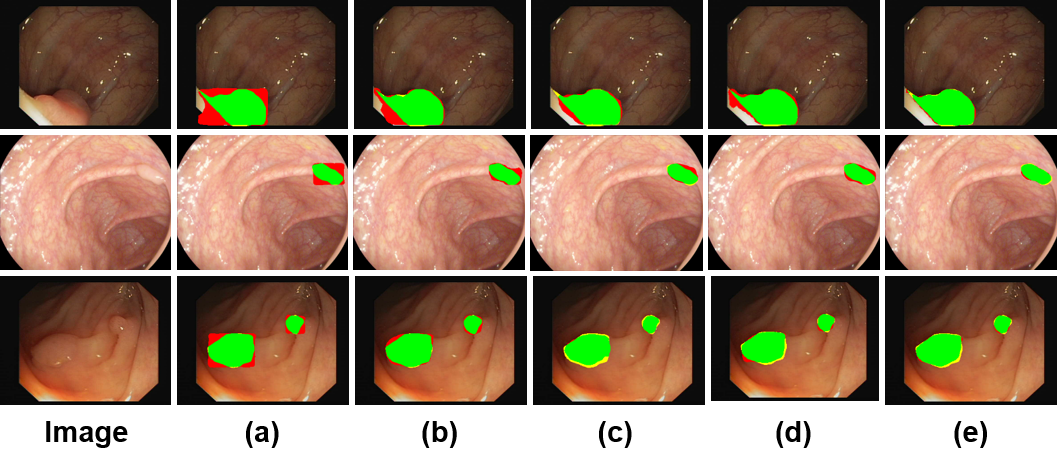}
	\caption{Three visualization results of different variants. (a)-(e) correspond to the 1$^{st}$ to the 5$^{th}$ rows in Table~\ref{table3}, respectively. Green: true positives; Yellow: false negatives; Red: false positives. }
	\label{ablation_study}
\end{figure}

The three loss terms, $\mathcal{L}_{ibox}$, $\mathcal{L}_{cla}$, $\mathcal{L}_{px}$ in Eq.~(\ref{eq7}), connect to three key factors affecting the performance, \ie, the improved box-dice, contrastive latent-anchors and teacher-provided masks.
To verify their effectiveness, following the dataset setting in Sec.~\ref{com_box}, we train three variants of IBoxCLA by disabling $\mathcal{L}_{cla}$ and/or $\mathcal{L}_{px}$, and a baseline method directly supervised by the box-filled masks.
The segmentation results are presented in Table~\ref{table3} and Fig.~\ref{ablation_study} both quantitatively and qualitatively.
Based on these results, three key conclusions can be made as follows:
\begin{table}[t]\small
\centering
\caption{Ablation study on effectiveness of the three loss terms.}
\scalebox{0.9}
{
	\begin{threeparttable}
\begin{tabular}{ccccccc}
	\toprule
 % \multirow{2}{*}{Constraints} &\multicolumn{3}{c}{Detection}
	 \multirow{2}{*}{$\mathcal{L}_{ibox}$} & \multirow{2}{*}{$\mathcal{L}_{cla}$} & \multirow{2}{*}{$\mathcal{L}_{px}$} & \multicolumn{2}{c}{Overall} & \multicolumn{2}{c}{$p$-values} \\
  &&&mDice &mIoU &mDice &mIoU \\
	\midrule
	\color{red}{\XSolidBrush} &\color{red}{\XSolidBrush} &\color{red}{\XSolidBrush} & 0.718 &0.593 & - & -\\
	\color{green}{\Checkmark} &\color{red}{\XSolidBrush}&\color{red}{\XSolidBrush}& 0.779 &0.690 &1.44E-28 &6.19E-60 \\
	\color{green}{\Checkmark} & \color{green}{\Checkmark} &\color{red}{\XSolidBrush} & 0.813 &0.727 &1.08E-12& 5.59E-13\\
	\color{green}{\Checkmark} &\color{red}{\XSolidBrush} &\color{green}{\Checkmark}  & 0.789 &0.708 &2.32E-2& 1.83E-2\\
	\color{green}{\Checkmark} & \color{green}{\Checkmark} & \color{green}{\Checkmark} & {0.824} &{0.746} &7.69E-11& 9.87E-14\\
	\bottomrule
\end{tabular}
\begin{tablenotes}
	\item[] The statistical test is performed on every two adjacent rows.
\end{tablenotes}
\end{threeparttable}
}
\label{table3}

\end{table}

(1) By comparing the first two rows in Table~\ref{table3} and Fig.~\ref{ablation_study}(a)-(b), the baseline predicts shape-collapsed masks, while IBox permits the model to learn shapes freely, leading to a significant increase of both mDice and mIoU on overall samples by $8.5\%$ and $16.4\%$, respectively ($p<0.001$ for both metrics).
But the errors around boundaries still occur for the variant using IBox only. 

(2) By comparing the middle three rows in Table~\ref{table3}, and Fig.~\ref{ablation_study}(b)-(d), both latent anchors and teacher-provided masks are effective, and each of them can facilitate the model to capture more reasonable shape boundaries.

(3) By comparing the last three rows in Table~\ref{table3}, and Fig.~\ref{ablation_study}(c)-(e), the auxiliary constraints at both feature-level and pixel-level are not mutually excluded, and the combination of them brings a significant improvement compared to IBox only, by an increase of mDice and mIoU on overall samples by $5.8\%$ and $8.1\%$, respectively ($p<0.001$ for both metrics).
Thus, IBoxCLA produces the most distinct polyp boundaries.

\begin{table}[t]\small
	\centering
	\caption{Comparison results of model's performance for polyp detection and delineation under the different constraints.}
	\resizebox{\linewidth}{!}{
		\begin{threeparttable}
		\begin{tabular}{p{3.3cm}p{0.9cm}<{\centering}p{0.9cm}<{\centering}p{0.9cm}<{\centering}p{1.5cm}<{\centering}}%{lcccccc}
			\toprule
			\multirow{2}{*}{Constraints} &\multicolumn{3}{c}{Detection} & Delineation \\
			\cmidrule{2-5}
			~&P $\uparrow$ &R $\uparrow$ &F1 $\uparrow$ &HD (px) $\downarrow$\\
			\midrule
			WeakPolyp & 0.771 &0.788 &0.777 &2.993\\
			\midrule
			Regular box-dice only &0.774 &0.828 &0.798 &5.579 \\
			$+$ Shape-decoupling &0.736 &0.801 &0.765 &3.110 \\
			$+$ Conf.-region swapping &0.763 &0.806 &0.784 &3.012 \\
			$+$ Cont. latent-anchors &{0.841} &{0.829} &{0.833} &{2.810}  \\
			\bottomrule
		\end{tabular}
	\begin{tablenotes}
		\item[] $\uparrow$ indicates that the higher the values, the better the accuracy for the metric. $\downarrow$ is the opposite.
	\end{tablenotes}
	\end{threeparttable}
	}
	\label{table4}
\end{table}

\subsubsection{Qualitative effectiveness of CLA in feature space}
\begin{figure}[t]
	\centering
	\includegraphics[width=\linewidth]{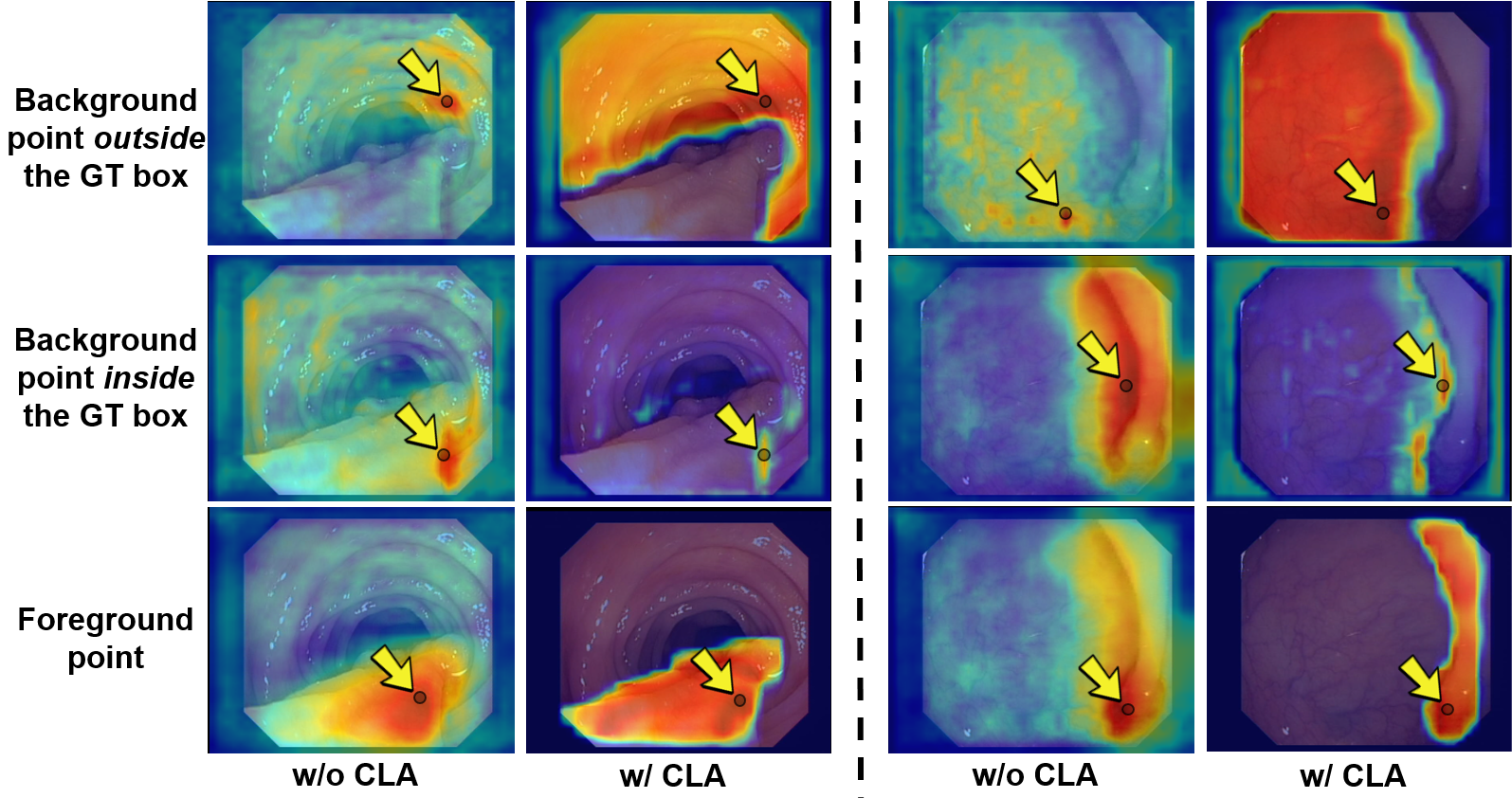}
	\caption{Similarity map of two examples w/o and w/ CLA in the training. Entries in the map are the cosine similarities between the feature vectors and a self-query indicated by the yellow arrow. The red color means high similarity and the blue means low similarity.}
	\label{ablation_study_CLA}
\end{figure}
To further verify our proposed CLA, we use the fused multi-scale feature map to calculate one-to-all cosine similarity maps leveraging three query feature vectors at different positions, \ie, background outside the GT box, background inside the GT box, and foreground.
Fig.~\ref{ablation_study_CLA} shows the similarity maps of two examples before and after adding CLA in the training.

As can be seen, before using CLA, each query point only activates small nearby areas.
Moreover, the activated areas by the in-box background query point percolate through the boundaries.
By using CLA, almost the entire regions of background and polyp respond to the corresponding query points.
And the activated areas by use of the in-box background query point obviously not exceed boundaries.
This comparison demonstrates that the feature contrast between polyp and background is majorly enhanced by CLA, and the features from the same category cluster tightly, which undoubtedly can facilitate the model to distinguish clearer boundaries.

\subsubsection{Performance on polyp detection/delineation}
\label{sec. ablation_decouple}
In this part, we reveal how the model behaves on learning locations/sizes (detection task) and shapes (delineation task) under individual constraints.
Following the dataset setting in Sec.~\ref{com_box}, we start from the baseline model trained with the regular box-dice constraint only, and add the shape-decoupling, confusion-region swapping and contrastive latent anchors, gradually.
Table~\ref{table4} shows the results under each constraint setting.
We also include WeakPolyp as a comparison.

The model in the $2^{nd}$ row achieves a clearly better performance on localizing polyps compared to WeakPolyp especially in terms of recall and F1 values, but unfortunately shows worst polyp delineation performance.
This verifies that the simplistic box-filled masks indeed contain sufficient guidance for capturing accurate polyp positions and sizes, but also mis-guidance hurting the shape learning. 

With the shape-decoupling added, HD dramatically drops (the $2^{nd}$ vs. $3^{rd}$ row), because the shape learning is set free from the dice constraint, and can be promoted by the unambiguous background features outside boxes.
However, the degraded detection performance also can be observed.
This is because of the caused confusion regions in the preliminary proxy map, leading to partial misdirection in the constraint.
With the confusion-region swapping added, the detection performance restores to the previous level (see the $4^{th}$ row), which means the confusion-region swapping is an indispensable step in the proxy map construction.

With the addition of contrastive latent anchors, the model learns more distinct features between the polyp and background regions, resulting in a significant improvement on polyp delineation, and in turn helping suppress false detection and recall more polyps as shown in the last row.
 
\subsection{Learning Efficiency under Mix-supervisions}
\label{sec. learning eff}
\begin{figure}[t]
    \centering
    \includegraphics[width=\linewidth]{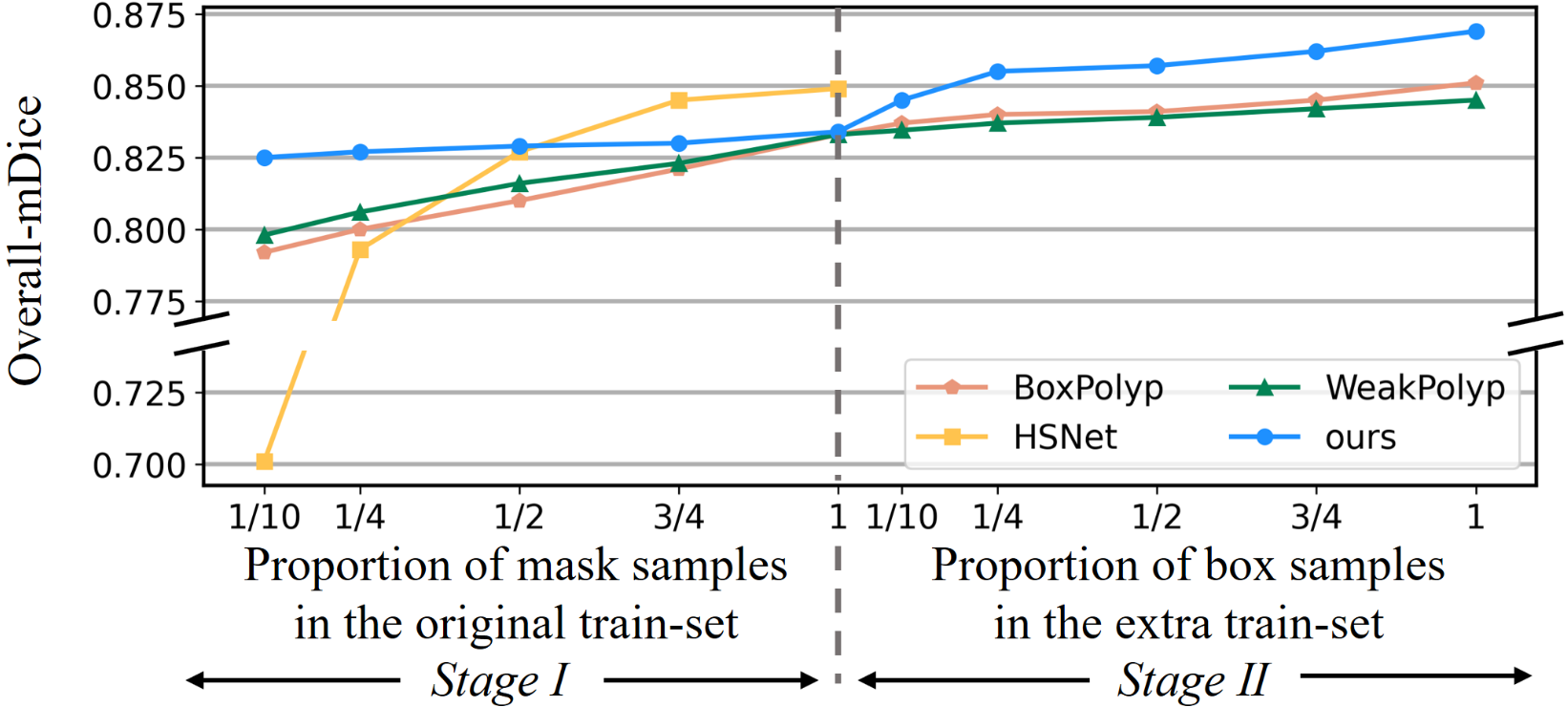}
    \caption{Overall-mDice of different methods which are trained on different proportions of train-set.}
    \label{Learning Efficiency}
\end{figure}
The value of this work lies in providing an accurate and efficient approach to absorb a large number of cheap annotated data, to further improve the segmentation performance in clinical usage. In this part, we conduct a two-stage experiment to provide a more in-depth analysis. Specifically, in stage \uppercase\expandafter{\romannumeral1}, only the original train-set (1,450 samples from ClincDB and Kvasir-SEG) is used for training, and we start by using $10\%$ samples as mask-annotated and the rest as box-annotated for mix-supervised training. We then gradually increase the proportion of samples as mask-labeled (the box-annotated decreases accordingly) until all samples become mask-annotated in the training. In stage \uppercase\expandafter{\romannumeral2}, we continue the training in the first stage but additionally add box-annotated samples from the extra train-set LDPolypVideo gradually.

Fig.~\ref{Learning Efficiency} illustrates the performance trends in terms of overall mDice on the test-set achieved by the best fully-supervised method, HSNet, and three top box-supervised methods, BoxPolyp, WeakPolyp and our proposed IBoxCLA.
As can be seen in the stage \uppercase\expandafter{\romannumeral1}, when the mask samples are less than $25\%$, all box-supervised methods outperform the HSNet, and IBoxCLA still surpasses it until the mask samples are over $50\%$.
Also, the performance trend of IBoxCLA is more stable and flat compared to those of WeakPolyp and BoxPolyp.
These imply that our method greatly maximize the use efficiency of box annotations.

In the stage \uppercase\expandafter{\romannumeral2}, HSNet has to stop since no more masks can be used, while the box-supervised methods are still improving.
From 0\% to 25\%, IBoxCLA gains a remarkably quicker improvement from $0.833$ to $0.856$ in terms of overall mDice ($p=0.002$), and exceeds HSNet by a non-trivial margin ($p=0.017$).
This well verifies the great practical value of IBoxCLA and its potential to further push the segmentation performance of SOTA architectures using other box-annotated samples.
\section{Conclusion and Future Work}
Leveraging coarser box-level annotations to train a finer model of pixel-level segmentation gains increasing attentions as a promising solution to alleviate the data shortage in the medical domain.
In this paper, we propose IBoxCLA, a new and intuitive box-supervised segmentation method for polyps.
IBoxCLA permits a handy and robust training with popular Dice loss original designed for fully-supervised methods.
It directly fills coarsely-annotated boxes as pseudo masks for Dice constraint, and introduces two novel learning fashions, i.e., IBox and CLA, to capture polyp locations/sizes and shapes accurately.
IBox aims to decouple the shape information to avoid misguidance carried by the box-filled masks, and CLA further enhances the boundaries using momentum and contrastive learning.
The comprehensive and extensive comparisons on five widely-employed polyp datasets and fourteen state-of-the-arts (SOTA) demonstrate that IBoxCLA outperforms the other box-supervised SOTAs significantly, and when using extra cheaper box-annotated samples, it is even superior over the fully-supervised SOTAs by a noticeable margin.
Ablation studies also verify the effectiveness of IBox and CLA both qualitatively and quantitatively, and the great application value of IBoxCLA in the realistic clinical practice.
Nevertheless, there still exit several challenges should be addressed in the future.
For example, the annotation noise that multiple polyps correspond to one big box or the boxes are too large/small will be explored and addressed.

\begin{small}
\vspace{.3in} \noindent \textbf{Data Availability:}
The images presented in this paper are derived from the following databases.
The ClinicDB, Kvasir-SEG, ColonDB, EndoScene, and ETIS datasets are publicly available at \url{https://github.com/DengPingFan/PraNet}; The LDPolypVideo dataset is publicy available at \url{https://github.com/dashishi/LDPolypVideo-Benchmark}.

\vspace{.3in} \noindent \textbf{Competing Interests:}
The authors declare that they have no competing financial or non-financial
interests related to this work.

\vspace{.3in} \noindent \textbf{Authors’ Contributions:}
% Qiang Hu and Ying Chen are co-first authors who developed the core ideas, wrote the draft, revised the manuscript, and conducted experiments. Zhiwei Wang is corresponding author who was in part of idea polishing, revised the manuscript and supervised the experiments. Qiang Li conducted partial experiments and drafted the experiment part. Hongkuan Shi is industrial co-author who gave technical supports, supervised the experiments and helped to review the manuscript.
% For research articles authors may use CRediT taxonomy:
Conceptualization: Qiang Hu, Zhiwei Wang;
Methodology: Qiang Hu, Ying Chen, Zhiwei Wang; 
Formal analysis and investigation: Qiang Hu, Ying Chen; 
Writing - original draft preparation: Qiang Hu, Ying Chen; 
Writing - review and editing: Hongkuan Shi, Qiang Li, Zhiwei Wang; 
Funding acquisition: Qiang Li, Zhiwei Wang; 
Resources: Hongkuan Shi, Qiang Li, Zhiwei Wang; 
Supervision: Qiang Li, Zhiwei Wang.
% %
% For review articles where discrete statements are less applicable a 
% statement should be included who had the idea for the article, 
% who performed the literature search and data analysis, 
% and who drafted and/or critically revised the work.

\end{small}

% acknowledgments part
\begin{acknowledgements}
This work was supported in part by the National Natural Science Foundation of China (Grant No.62576145), research grants from Wuhan United Imaging Healthcare Surgical Technology Co., Ltd.
\end{acknowledgements}

% BibTeX from reference.bib
%\bibliographystyle{sn-apacite}
\bibliographystyle{unsrt}
\bibliography{reference}

@string(MICCAI= "{Med. Image. Comput. Comput. Assist. Interv.}")

@string(TOG   = "{ACM Trans. Graph.}")

@article{sung2021global,
  title={Global cancer statistics 2020: GLOBOCAN estimates of incidence and mortality worldwide for 36 cancers in 185 countries},
  author={Sung, Hyuna and Ferlay, Jacques and Siegel, Rebecca L and Laversanne, Mathieu and Soerjomataram, Isabelle and Jemal, Ahmedin and Bray, Freddie},
  journal={CA: a cancer journal for clinicians},
  volume={71},
  number={3},
  pages={209--249},
  year={2021},
  publisher={Wiley Online Library}
}

@article{haggar2009colorectal,
  title={Colorectal cancer epidemiology: incidence, mortality, survival, and risk factors},
  author={Haggar, Fatima A and Boushey, Robin P},
  journal={Clinics in colon and rectal surgery},
  volume={22},
  number={04},
  pages={191--197},
  year={2009},
  publisher={{\copyright} Thieme Medical Publishers}
}

@inproceedings{tian2021boxinst,
  title={Boxinst: High-performance instance segmentation with box annotations},
  author={Tian, Zhi and Shen, Chunhua and Wang, Xinlong and Chen, Hao},
  booktitle={Proceedings of the IEEE/CVF Conference on Computer Vision and Pattern Recognition},
  pages={5443--5452},
  year={2021}
}

@inproceedings{ronneberger2015u,
  title={U-net: Convolutional networks for biomedical image segmentation},
  author={Ronneberger, Olaf and Fischer, Philipp and Brox, Thomas},
  booktitle={International Conference on Medical image computing and computer-assisted intervention},
  pages={234--241},
  year={2015},
  organization={Springer}
}

@inproceedings{fan2020pranet,
  title={Pranet: Parallel reverse attention network for polyp segmentation},
  author={Fan, Deng-Ping and Ji, Ge-Peng and Zhou, Tao and Chen, Geng and Fu, Huazhu and Shen, Jianbing and Shao, Ling},
  booktitle={International conference on medical image computing and computer-assisted intervention},
  pages={263--273},
  year={2020},
  organization={Springer}
}

@inproceedings{wei2021shallow,
  title={Shallow attention network for polyp segmentation},
  author={Wei, Jun and Hu, Yiwen and Zhang, Ruimao and Li, Zhen and Zhou, S Kevin and Cui, Shuguang},
  booktitle={International Conference on Medical Image Computing and Computer-Assisted Intervention},
  pages={699--708},
  year={2021},
  organization={Springer}
}

@article{dong2021polyp,
  title={Polyp-pvt: Polyp segmentation with pyramid vision transformers},
  author={Dong, Bo and Wang, Wenhai and Fan, Deng-Ping and Li, Jinpeng and Fu, Huazhu and Shao, Ling},
  journal={arXiv preprint arXiv:2108.06932},
  year={2021}
}

@inproceedings{kulharia2020box2seg,
  title={Box2seg: Attention weighted loss and discriminative feature learning for weakly supervised segmentation},
  author={Kulharia, Viveka and Chandra, Siddhartha and Agrawal, Amit and Torr, Philip and Tyagi, Ambrish},
  booktitle={European Conference on Computer Vision},
  pages={290--308},
  year={2020},
  organization={Springer}
}

@article{rother2004grabcut,
  title={" GrabCut" interactive foreground extraction using iterated graph cuts},
  author={Rother, Carsten and Kolmogorov, Vladimir and Blake, Andrew},
  journal={ACM transactions on graphics (TOG)},
  volume={23},
  number={3},
  pages={309--314},
  year={2004},
  publisher={ACM New York, NY, USA}
}

@inproceedings{lan2021discobox,
  title={Discobox: Weakly supervised instance segmentation and semantic correspondence from box supervision},
  author={Lan, Shiyi and Yu, Zhiding and Choy, Christopher and Radhakrishnan, Subhashree and Liu, Guilin and Zhu, Yuke and Davis, Larry S and Anandkumar, Anima},
  booktitle={Proceedings of the IEEE/CVF International Conference on Computer Vision},
  pages={3406--3416},
  year={2021}
}

@inproceedings{li2022box,
  title={Box-supervised instance segmentation with level set evolution},
  author={Li, Wentong and Liu, Wenyu and Zhu, Jianke and Cui, Miaomiao and Hua, Xian-Sheng and Zhang, Lei},
  booktitle={European Conference on Computer Vision},
  pages={1--18},
  year={2022},
  organization={Springer}
}

@inproceedings{fan2022self,
  title={Self-support few-shot semantic segmentation},
  author={Fan, Qi and Pei, Wenjie and Tai, Yu-Wing and Tang, Chi-Keung},
  booktitle={European Conference on Computer Vision},
  pages={701--719},
  year={2022},
  organization={Springer}
}

@article{bernal2015wm,
  title={WM-DOVA maps for accurate polyp highlighting in colonoscopy: Validation vs. saliency maps from physicians},
  author={Bernal, Jorge and S{\'a}nchez, F Javier and Fern{\'a}ndez-Esparrach, Gloria and Gil, Debora and Rodr{\'\i}guez, Cristina and Vilari{\~n}o, Fernando},
  journal={Computerized medical imaging and graphics},
  volume={43},
  pages={99--111},
  year={2015},
  publisher={Elsevier}
}

@inproceedings{jha2020kvasir,
  title={Kvasir-seg: A segmented polyp dataset},
  author={Jha, Debesh and Smedsrud, Pia H and Riegler, Michael A and Halvorsen, P{\aa}l and Lange, Thomas de and Johansen, Dag and Johansen, H{\aa}vard D},
  booktitle={International Conference on Multimedia Modeling},
  pages={451--462},
  year={2020},
  organization={Springer}
}

@article{tajbakhsh2015automated,
  title={Automated polyp detection in colonoscopy videos using shape and context information},
  author={Tajbakhsh, Nima and Gurudu, Suryakanth R and Liang, Jianming},
  journal={IEEE transactions on medical imaging},
  volume={35},
  number={2},
  pages={630--644},
  year={2015},
  publisher={IEEE}
}

@article{vazquez2017benchmark,
  title={A benchmark for endoluminal scene segmentation of colonoscopy images},
  author={V{\'a}zquez, David and Bernal, Jorge and S{\'a}nchez, F Javier and Fern{\'a}ndez-Esparrach, Gloria and L{\'o}pez, Antonio M and Romero, Adriana and Drozdzal, Michal and Courville, Aaron},
  journal={Journal of healthcare engineering},
  volume={2017},
  year={2017},
  publisher={Hindawi}
}

@article{silva2014toward,
  title={Toward embedded detection of polyps in wce images for early diagnosis of colorectal cancer},
  author={Silva, Juan and Histace, Aymeric and Romain, Olivier and Dray, Xavier and Granado, Bertrand},
  journal={International journal of computer assisted radiology and surgery},
  volume={9},
  number={2},
  pages={283--293},
  year={2014},
  publisher={Springer}
}

@inproceedings{fang2019selective,
  title={Selective feature aggregation network with area-boundary constraints for polyp segmentation},
  author={Fang, Yuqi and Chen, Cheng and Yuan, Yixuan and Tong, Kai-yu},
  booktitle={International Conference on Medical Image Computing and Computer-Assisted Intervention},
  pages={302--310},
  year={2019},
  organization={Springer}
}

@inproceedings{dai2015boxsup,
  title={Boxsup: Exploiting bounding boxes to supervise convolutional networks for semantic segmentation},
  author={Dai, Jifeng and He, Kaiming and Sun, Jian},
  booktitle={Proceedings of the IEEE international conference on computer vision},
  pages={1635--1643},
  year={2015}
}

@inproceedings{groger2022boxshrink,
  title={BoxShrink: From Bounding Boxes to Segmentation Masks},
  author={Gr{\"o}ger, Michael and Borisov, Vadim and Kasneci, Gjergji},
  booktitle={Medical Image Learning with Limited and Noisy Data: First International Workshop, MILLanD 2022, Held in Conjunction with MICCAI 2022, Singapore, September 22, 2022, Proceedings},
  pages={65--75},
  year={2022},
  organization={Springer}
}

@inproceedings{mahani2022bounding,
  title={Bounding Box Based Weakly Supervised Deep Convolutional Neural Network for Medical Image Segmentation Using an Uncertainty Guided and Spatially Constrained Loss},
  author={Mahani, Golnar K and Li, Ruizhe and Evangelou, Nikolaos and Sotiropolous, Stamatios and Morgan, Paul S and French, Andrew P and Chen, Xin},
  booktitle={2022 IEEE 19th International Symposium on Biomedical Imaging (ISBI)},
  pages={1--5},
  year={2022},
  organization={IEEE}
}

@inproceedings{wang2021bounding,
  title={Bounding box tightness prior for weakly supervised image segmentation},
  author={Wang, Juan and Xia, Bin},
  booktitle={Medical Image Computing and Computer Assisted Intervention--MICCAI 2021: 24th International Conference, Strasbourg, France, September 27--October 1, 2021, Proceedings, Part II},
  pages={526--536},
  year={2021},
  organization={Springer}
}

@inproceedings{cheng2023boxteacher,
  title={Boxteacher: Exploring high-quality pseudo labels for weakly supervised instance segmentation},
  author={Cheng, Tianheng and Wang, Xinggang and Chen, Shaoyu and Zhang, Qian and Liu, Wenyu},
  booktitle={Proceedings of the IEEE/CVF Conference on Computer Vision and Pattern Recognition},
  pages={3145--3154},
  year={2023}
}

@inproceedings{zhao2021automatic,
  title={Automatic polyp segmentation via multi-scale subtraction network},
  author={Zhao, Xiaoqi and Zhang, Lihe and Lu, Huchuan},
  booktitle={Medical Image Computing and Computer Assisted Intervention--MICCAI 2021: 24th International Conference, Strasbourg, France, September 27--October 1, 2021, Proceedings, Part I 24},
  pages={120--130},
  year={2021},
  organization={Springer}
}

@article{hsu2019weakly,
  title={Weakly supervised instance segmentation using the bounding box tightness prior},
  author={Hsu, Cheng-Chun and Hsu, Kuang-Jui and Tsai, Chung-Chi and Lin, Yen-Yu and Chuang, Yung-Yu},
  journal={Advances in Neural Information Processing Systems},
  volume={32},
  year={2019}
}

@inproceedings{wei2022boxpolyp,
  title={BoxPolyp: Boost Generalized Polyp Segmentation Using Extra Coarse Bounding Box Annotations},
  author={Wei, Jun and Hu, Yiwen and Li, Guanbin and Cui, Shuguang and Kevin Zhou, S and Li, Zhen},
  booktitle={Medical Image Computing and Computer Assisted Intervention--MICCAI 2022: 25th International Conference, Singapore, September 18--22, 2022, Proceedings, Part III},
  pages={67--77},
  year={2022},
  organization={Springer}
}

@article{chen2022weakly,
  title={Weakly Supervised Polyp Segmentation in Colonoscopy Images Using Deep Neural Networks},
  author={Chen, Siwei and Urban, Gregor and Baldi, Pierre},
  journal={Journal of Imaging},
  volume={8},
  number={5},
  pages={121},
  year={2022},
  publisher={MDPI}
}

@article{achanta2012slic,
  title={SLIC superpixels compared to state-of-the-art superpixel methods},
  author={Achanta, Radhakrishna and Shaji, Appu and Smith, Kevin and Lucchi, Aurelien and Fua, Pascal and S{\"u}sstrunk, Sabine},
  journal={IEEE transactions on pattern analysis and machine intelligence},
  volume={34},
  number={11},
  pages={2274--2282},
  year={2012},
  publisher={IEEE}
}

@article{zhang2022hsnet,
  title={HSNet: A hybrid semantic network for polyp segmentation},
  author={Zhang, Wenchao and Fu, Chong and Zheng, Yu and Zhang, Fangyuan and Zhao, Yanli and Sham, Chiu-Wing},
  journal={Computers in biology and medicine},
  volume={150},
  pages={106173},
  year={2022},
  publisher={Elsevier}
}

@inproceedings{ma2021ldpolypvideo,
  title={LDPolypVideo benchmark: a large-scale colonoscopy video dataset of diverse polyps},
  author={Ma, Yiting and Chen, Xuejin and Cheng, Kai and Li, Yang and Sun, Bin},
  booktitle={Medical Image Computing and Computer Assisted Intervention--MICCAI 2021: 24th International Conference, Strasbourg, France, September 27--October 1, 2021, Proceedings, Part V 24},
  pages={387--396},
  year={2021},
  organization={Springer}
}

@article{wei2023weakpolyp,
  title={WeakPolyp: You Only Look Bounding Box for Polyp Segmentation},
  author={Wei, Jun and Hu, Yiwen and Cui, Shuguang and Zhou, S Kevin and Li, Zhen},
  journal={arXiv preprint arXiv:2307.10912},
  year={2023}
}

@inproceedings{wu2019cascaded,
  title={Cascaded partial decoder for fast and accurate salient object detection},
  author={Wu, Zhe and Su, Li and Huang, Qingming},
  booktitle={Proceedings of the IEEE/CVF conference on computer vision and pattern recognition},
  pages={3907--3916},
  year={2019}
}

@inproceedings{zhou2018unet++,
  title={Unet++: A nested u-net architecture for medical image segmentation},
  author={Zhou, Zongwei and Rahman Siddiquee, Md Mahfuzur and Tajbakhsh, Nima and Liang, Jianming},
  booktitle={Deep Learning in Medical Image Analysis and Multimodal Learning for Clinical Decision Support: 4th International Workshop, DLMIA 2018, and 8th International Workshop, ML-CDS 2018, Held in Conjunction with MICCAI 2018, Granada, Spain, September 20, 2018, Proceedings 4},
  pages={3--11},
  year={2018},
  organization={Springer}
}

@article{chen2021transunet,
  title={Transunet: Transformers make strong encoders for medical image segmentation},
  author={Chen, Jieneng and Lu, Yongyi and Yu, Qihang and Luo, Xiangde and Adeli, Ehsan and Wang, Yan and Lu, Le and Yuille, Alan L and Zhou, Yuyin},
  journal={arXiv preprint arXiv:2102.04306},
  year={2021}
}

@inproceedings{cao2022swin,
  title={Swin-unet: Unet-like pure transformer for medical image segmentation},
  author={Cao, Hu and Wang, Yueyue and Chen, Joy and Jiang, Dongsheng and Zhang, Xiaopeng and Tian, Qi and Wang, Manning},
  booktitle={European conference on computer vision},
  pages={205--218},
  year={2022},
  organization={Springer}
}

@inproceedings{milletari2016v,
  title={V-net: Fully convolutional neural networks for volumetric medical image segmentation},
  author={Milletari, Fausto and Navab, Nassir and Ahmadi, Seyed-Ahmad},
  booktitle={2016 fourth international conference on 3D vision (3DV)},
  pages={565--571},
  year={2016},
  organization={Ieee}
}

@inproceedings{mamalet2012simplifying,
  title={Simplifying convnets for fast learning},
  author={Mamalet, Franck and Garcia, Christophe},
  booktitle={International Conference on Artificial Neural Networks},
  pages={58--65},
  year={2012},
  organization={Springer}
}

@article{chen2017deeplab,
  title={Deeplab: Semantic image segmentation with deep convolutional nets, atrous convolution, and fully connected crfs},
  author={Chen, Liang-Chieh and Papandreou, George and Kokkinos, Iasonas and Murphy, Kevin and Yuille, Alan L},
  journal={IEEE transactions on pattern analysis and machine intelligence},
  volume={40},
  number={4},
  pages={834--848},
  year={2017},
  publisher={IEEE}
}

@inproceedings{schroff2015facenet,
  title={Facenet: A unified embedding for face recognition and clustering},
  author={Schroff, Florian and Kalenichenko, Dmitry and Philbin, James},
  booktitle={Proceedings of the IEEE conference on computer vision and pattern recognition},
  pages={815--823},
  year={2015}
}

@inproceedings{he2020momentum,
  title={Momentum contrast for unsupervised visual representation learning},
  author={He, Kaiming and Fan, Haoqi and Wu, Yuxin and Xie, Saining and Girshick, Ross},
  booktitle={Proceedings of the IEEE/CVF conference on computer vision and pattern recognition},
  pages={9729--9738},
  year={2020}
}

@article{paszke2019pytorch,
  title={Pytorch: An imperative style, high-performance deep learning library},
  author={Paszke, Adam and Gross, Sam and Massa, Francisco and Lerer, Adam and Bradbury, James and Chanan, Gregory and Killeen, Trevor and Lin, Zeming and Gimelshein, Natalia and Antiga, Luca and others},
  journal={Advances in neural information processing systems},
  volume={32},
  year={2019}
}

@article{loshchilov2017decoupled,
  title={Decoupled weight decay regularization},
  author={Loshchilov, Ilya and Hutter, Frank},
  journal={arXiv preprint arXiv:1711.05101},
  year={2017}
}

@article{vaswani2017attention,
  title={Attention is all you need},
  author={Vaswani, Ashish and Shazeer, Noam and Parmar, Niki and Uszkoreit, Jakob and Jones, Llion and Gomez, Aidan N and Kaiser, {\L}ukasz and Polosukhin, Illia},
  journal={Advances in neural information processing systems},
  volume={30},
  year={2017}
}

@article{pont2016multiscale,
  title={Multiscale combinatorial grouping for image segmentation and object proposal generation},
  author={Pont-Tuset, Jordi and Arbelaez, Pablo and Barron, Jonathan T and Marques, Ferran and Malik, Jitendra},
  journal={IEEE transactions on pattern analysis and machine intelligence},
  volume={39},
  number={1},
  pages={128--140},
  year={2016},
  publisher={IEEE}
}

@inproceedings{lee2022mpvit,
  title={Mpvit: Multi-path vision transformer for dense prediction},
  author={Lee, Youngwan and Kim, Jonghee and Willette, Jeffrey and Hwang, Sung Ju},
  booktitle={Proceedings of the IEEE/CVF Conference on Computer Vision and Pattern Recognition},
  pages={7287--7296},
  year={2022}
}

@inproceedings{yu2016unitbox,
  title={Unitbox: An advanced object detection network},
  author={Yu, Jiahui and Jiang, Yuning and Wang, Zhangyang and Cao, Zhimin and Huang, Thomas},
  booktitle={Proceedings of the 24th ACM international conference on Multimedia},
  pages={516--520},
  year={2016}
}

@article{student1908probable,
  title={The probable error of a mean},
  author={Student},
  journal={Biometrika},
  volume={6},
  number={1},
  pages={1--25},
  year={1908},
  publisher={Oxford University Press}
}

@article{smedsrud2021kvasir,
  title={Kvasir-Capsule, a video capsule endoscopy dataset},
  author={Smedsrud, Pia H and Thambawita, Vajira and Hicks, Steven A and Gjestang, Henrik and Nedrejord, Oda Olsen and N{\ae}ss, Espen and Borgli, Hanna and Jha, Debesh and Berstad, Tor Jan Derek and Eskeland, Sigrun L and others},
  journal={Scientific Data},
  volume={8},
  number={1},
  pages={142},
  year={2021},
  publisher={Nature Publishing Group UK London}
}

@article{li2021colonoscopy,
  title={Colonoscopy polyp detection and classification: Dataset creation and comparative evaluations},
  author={Li, Kaidong and Fathan, Mohammad I and Patel, Krushi and Zhang, Tianxiao and Zhong, Cuncong and Bansal, Ajay and Rastogi, Amit and Wang, Jean S and Wang, Guanghui},
  journal={Plos one},
  volume={16},
  number={8},
  pages={e0255809},
  year={2021},
  publisher={Public Library of Science San Francisco, CA USA}
}

\end{document}